\documentclass[letterpaper,journal]{IEEEtran}
\usepackage{amsmath,amsfonts}
\usepackage{algorithmic}
\usepackage{algorithm}
\usepackage{array}
\usepackage[caption=false,font=normalsize,labelfont=sf,textfont=sf]{subfig}
\usepackage{textcomp}
\usepackage{stfloats}
\usepackage{url}
\usepackage{verbatim}
\usepackage{graphicx}
\usepackage{cite}

\usepackage{mdwmath}
\usepackage{mdwtab}
\usepackage{eqparbox}
\usepackage{enumerate}
\usepackage{hyperref} 
\usepackage[all]{hypcap}
\usepackage{booktabs}
\usepackage{multirow}
\usepackage{makecell}
\usepackage{color}
\usepackage{wrapfig}
\usepackage{diagbox}

\usepackage{xcolor}
\newif\ifShowChanges
\ShowChangesfalse % black for final

\ifShowChanges
  \DeclareRobustCommand{\rev}[1]{\textcolor{black}{#1}}%
\else
  \DeclareRobustCommand{\rev}[1]{#1}%
\fi

\ifShowChanges
  \DeclareRobustCommand{\revtwo}[1]{\textcolor{red}{#1}}%
\else
  \DeclareRobustCommand{\revtwo}[1]{#1}%
\fi

\begin{document}

\title{Tracking Large-scale Shared Bikes with Inertial Motion Learning in GNSS Blocked Environments}

\author{Feng~Liu, Kejia~Li, Zhiwei~Yang, Chunwei~Yang, Qun~Li, Guobin~Wu, Qiang~Ni, Ruipeng~Gao$^{*}$
        % ~\IEEEmembership{Staff,~IEEE,}
        % <-this % stops a space
\thanks{This paper is supported in part by the Natural Science Foundation of
China grant No. 62572041, Beijing Nova Program 20230484263 and 20240484607, and DiDi Research Collaboration Plan. Corresponding author: Ruipeng Gao.
}% <-this % stops a space

\thanks{Feng~Liu and Kejia~Li are with the School of Software Engineering, Beijing Jiaotong University, Beijing 100044, China. Email: \{liu.feng, kejiali\}@bjtu.edu.cn.}
\thanks{Ruipeng~Gao is with the School of Cyberspace Science and Technology, Beijing Jiaotong University, Beijing 100044, China. Email: rpgao@bjtu.edu.cn.}
\thanks{Zhiwei~Yang, Chunwei~Yang, Qun~Li and Guobin~Wu are with DiDi, Beijing 100000, China. Email: \{yangzhiwei, chunweiyang, liquntracy, wuguobin\}@didiglobal.com.}
\thanks{Qiang~Ni is with the School of Computing and Communications, Lancaster University, Lancaster, LA1 4WA, UK. E-mail: q.ni@lancaster.ac.uk.}
}

% The paper headers
\markboth{Journal of \LaTeX\ Class Files,~Vol.~14, No.~8, August~2021}%
{Shell \MakeLowercase{\textit{et al.}}: A Sample Article Using IEEEtran.cls for IEEE Journals}

% \IEEEpubid{0000--0000/00\$00.00~\copyright~2021 IEEE}
% Remember, if you use this you must call \IEEEpubidadjcol in the second
% column for its text to clear the IEEEpubid mark.

\maketitle

\begin{abstract}

Although Global Navigation Satellite Systems (GNSS) provide a general solution for bike tracking outdoors, there still exist complex riding environments where only inertial navigation systems work, such as urban canyons.
Despite decades of research, localization using only low-cost inertial sensors still faces challenges such as cumulative drifts and poor robustness caused by filtering methods.   Furthermore, sensors such as visual and LiDAR could provide reliable measurements, but they are not suitable for large-scale deployment.
In this paper, we propose an inertial tracking framework that integrates bicycle mechanical constraints with a mixture-of-experts model.
Specifically, we leverage multiple expert modules to capture shared representations and weight them through the gating mechanism, thus improving multi-task learning performance and enabling uncertainty-aware trajectory estimation.
%we explore interpretable motion variables, leverage adaptive gating to enhance representation learning, and enable uncertainty-aware trajectory estimation.
Furthermore, based on the mechanical transmission between the pedal and the rear wheel of a bike, we explore the intrinsic relationship between the rider's periodic pedalling behaviors and acceleration variations, and convert such patterns into bike's wheel speed for dynamic calibration.
Experiments with real-world riding data from shared bikes of the DiDi ride-hailing platform demonstrate that our system improves the accuracy of baselines by at least $12\%$, with wheel speed errors below $0.5 m/s$ at 95-percentile.

\end{abstract}


\section{Introduction}\label{sec:intro}

\begin{figure}[t]
\label{fig_complex_area}
    \centering
    \begin{minipage}{1.0\linewidth}
    % \vspace{3pt}
    \centerline{\includegraphics[width=\textwidth]{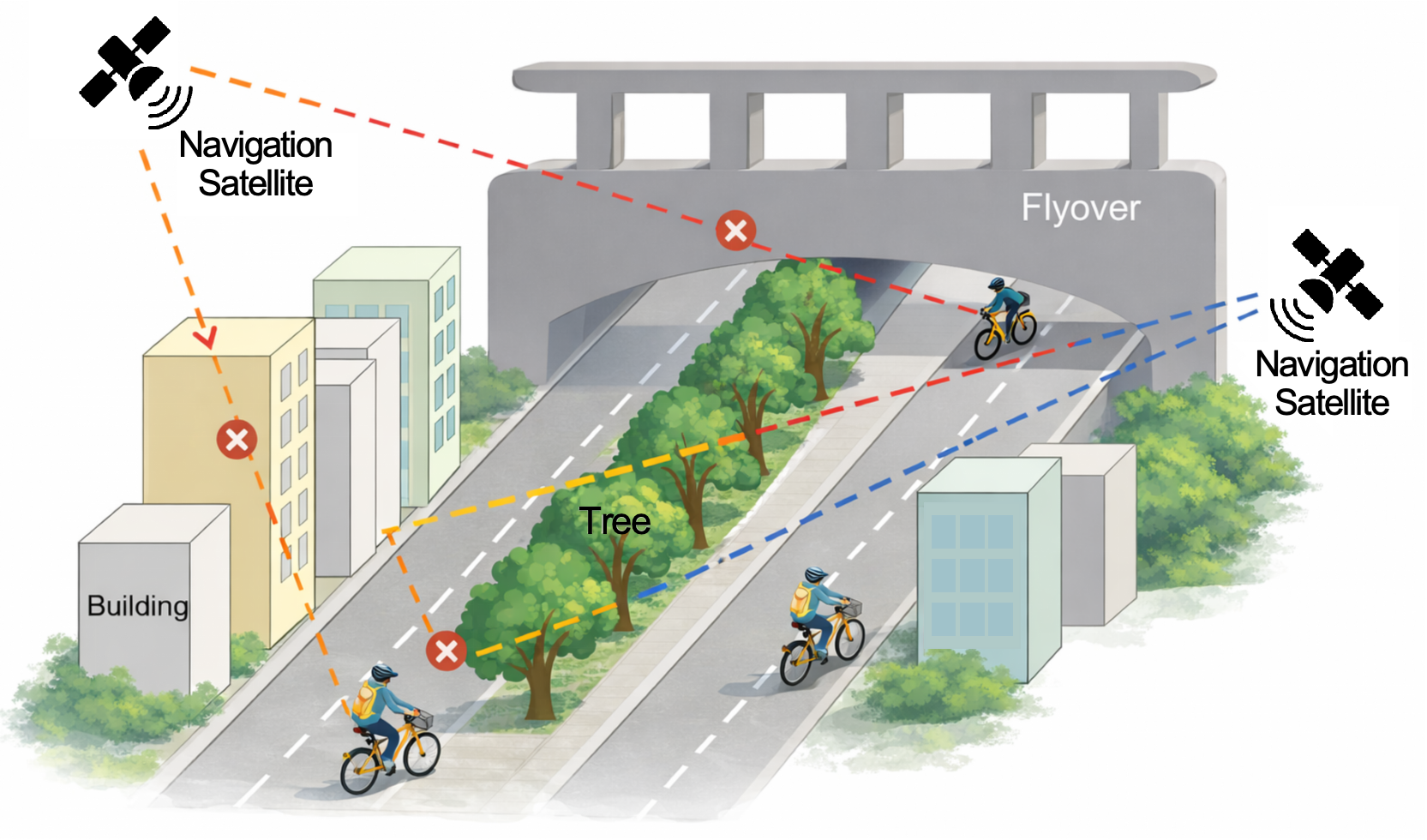}}
    \end{minipage}
    \caption{Satellite signals may be blocked or even unavailable in many bike-riding environments, e.g., dense building complexes, multi-level flyovers, and tree-lined roads.}
\end{figure}

\IEEEPARstart{S}{hared} bikes have become a key solution to the ``last mile'' transportation problem in urban cities due to the flexibility, low cost, and sustainability. 
Thanks to the prevalence of Global Navigation Satellite Systems (GNSS) and embedded devices, we can access and track shared bikes in most urban areas.
Such information is essential for bicycle location-based services, including bike retrieval, route planning, and dispatch management.
However, whenever we ride through satellite blocked areas (Figure~\ref{fig_complex_area}), such as dense building complexes, multi-level flyovers, and tree-lined roads, we lose the location awareness. 
%Not only does it lead to cumulative trajectory drift, but it also reduces the observability of motion states.

Enabling stable bike tracking in GNSS blocked environments is crucial for users to lock the device and pay the bill, as well as for ride-hailing platforms to dispatch and manage millions of shared bikes.
For example, when a user returns a shared bike, the platform must determine whether it is parked in correct region.
Put simply, shared bikes need to be always aware of their spatial position in the environment.

However, tracking shared bicycles in scenarios without GNSS signals is far from straightforward.
Inertial Measurement Unit (IMU) provides high-frequency motion observations independent of the environment, but integration is vulnerable to severe drifts due to sensor noises, zero bias, and error accumulations.
Although the Integrated Navigation System (INS) employs algorithms such as the Kalman filter~\cite{kalman1960new} to fuse the absolute position from GNSS with the real-time motion from the IMU, thus suppressing inertial accumulation errors, it remains susceptible to failure in urban environments with signal blockages.
Mainstream indoor localization approaches leverage Radio-Frequency signals such as Wi-Fi~\cite{zhao2022gltc}, Bluetooth~\cite{10695776}, and Ultra-wideband~\cite{10260266}, which are cost-expensive to deploy widely and calibrate automatically for industry.
The low-cost, low-power embedded devices equipped on shared bikes constrain the types and quality of sensing modality, thus impedes the performance of real-time localization.

To address the issue of cumulative errors, some studies have proposed deep learning methods that directly regress velocity or displacement from inertial sequences, to avoid double integration, such as IONet~\cite{Chen_Lu_Markham_Trigoni_2018}, RoNIN~\cite{9196860}, and LLIO~\cite{wang2022llio}. 
% However, these approaches typically employ one module to estimate multiple variables simultaneously, which leads to ineffectiveness in capturing the differences and correlations between distinct variables. 
\rev{However, these approaches typically rely on common latent features, with fixed task-specific outputs, which limits their capacity to capture unique yet correlated relationships between different tasks. 
In contrast, we further decompose the task, treating motion changes and their corresponding residuals and uncertainties as interrelated yet distinct tasks.} 
Alternatively, some methods design specific error constraints based on domain-specific knowledge, using techniques such as stationary detection~\cite{10137733} and map matching~\cite{xia2023integrated} to correct errors. 
Nevertheless, they largely rely on specific assumptions or additional information, which are not suitable for large-scale deployment.

% In this paper, we only leverages inertial data collected by the embedded IMU to track shared bikes in real time, and avoid reliance on additional sensors such as the energy hungry GNSS. 
In order to improve the localization performance of an inertial-only dead-reckoning system, we also need to calibrate cumulative errors from long-term drifts.
Such inertia-only solutions face multiple challenges in practice. First, the low-cost IMUs embedded in shared bikes are plagued by severe noises, causing unbounded positioning errors in double integrations. 
Second, inertial dead-reckoning inherently involves uncertainty accumulation, thus we need to calculate the uncertainty of such estimates.
Finally, without GNSS signals, shared bikes lack effective measurements for calibration, thus resulting in accumulated errors during tracking.
%Under inertial-only conditions, introducing effective observations is also critical for promoting practical use.

% Our main intuition is to discover periodic fluctuations in inertia readings during riding. Specially, we identify obvious periodic patterns in acceleration measurements, which can approximate the bike wheel speed. This observation enables calibration of the bike speed estimates in real time, thus it is referred to as ``pseudo wheel speed''. In addition, we explore a lightweight inertial learning model to infer the real-time position of shared bikes regardless of GNSS signals. 

To address such challenges, our main intuition is to discover  periodic fluctuations in inertial readings during riding, which are influenced by the mechanical transmission between the bike pedal and the rear wheel, \rev{and are similar to the periodic biomechanical patterns reported in \cite{OBRYAN201411}.} 
Specifically, we identify such periodic patterns in forwarding accelerations, which approximate the bicycle wheel speed. 
This observation enables calibration of the bike's speed estimates in real time, thus it is referred to as ``pseudo wheel speed''. 
In addition, we design a Mixture-of-Experts (MoE) based inertial learning model to capture potential correlations across different tasks, thus improving localization accuracy and robustness.

In sum, our contributions include:

\begin{itemize}
    \item We propose a Multi-Task Inertial Motion Network (MTIMNet) that leverages a mixture-of-experts architecture to learn bike's motion representations. We also estimate the prediction errors in real time to ensure interpretability for the service.
    % We devise a Multi-Task Inertial Motion Network (MTIMNet) that learns interpretable motion representations from IMU sequences and generates continuous motion trajectories for shared bikes. We also model the estimation confidence to provide corresponding industrial operations.
    \item We propose a kinematically inspired wheel speed estimation method. It detects periodic patterns in forwarding accelerations to directly derive bike's velocity, achieving dynamic calibration without relying on additional sensors or machine learning models.
    % We estimate the bike wheel speed based on rider's pedalling behaviors. It detects periodic patterns in accelerations, and directly derives the bike's pseudo-speed without relying on machine learning models.
    \item We implement an integrated prototype and conduct extensive evaluations on real-world riding data collected by a modern ride-hailing platform. Results demonstrate that our solution outperforms baselines, achieving at least a 12\% improvement in both Absolute Trajectory Error (ATE) and Absolute Yaw Error (AYE).
\end{itemize}

The rest of this paper is organized as follows. 
Section~\ref{sec:related} reviews related work in the field of localization.
Section~\ref{sec:preliminary} provides preliminary background. 
Section~\ref{sec:overview} presents an overview of our tracking framework. 
Section~\ref{sec:mtimnet} details the design of the Multi-Task Inertial Motion Network. 
Section~\ref{sec:wheel} analyzes the feasibility of the pseudo wheel speed estimation algorithm for shared bikes. 
Section~\ref{sec:evaluation} demonstrates extensive evaluation results. 
Section~\ref{sec:conclusion} concludes the paper.

\section{Related Work}\label{sec:related}

\subsection{Inertial navigation methods}

Classic Strapdown Inertial Navigation Systems (SINS) estimate pose by integrating IMU measurements, but accumulated sensor errors cause severe drift, especially for low-cost IMUs~\cite{chen2024deep}. 
To improve the robustness of localization, recent studies employ Deep Neural Networks (DNNs) to learn motion features and relieve errors.

On one hand, some studies leverage DNNs to calibrate inertial measurement errors. 
For example, \cite{chen2018improving} employed a convolutional network to denoise low-quality data using high-quality references. 
Calib-Net~\cite{li2022calib} utilized dilated convolutions to extract spatio-temporal features for gyroscope noise compensation. 
Similar works~\cite{nobre2019learning, brossard2020denoising, huang2022mems, qiu2024airimulearninguncertaintypropagation} show that DNNs can partially suppress inertial noise, but the performance of the models remains susceptible to variations in sensors or users.

% \cite{wang2020pedestrian} integrated velocity and noise estimation to enhance the extended Kalman filter (KF). 

% old
% On the other hand, many studies have established data-driven inertial motion models to mitigate the accumulation errors of SINS. 
% A first line of work directly learns local motion variables from IMU sequences, such as IONet~\cite{Chen_Lu_Markham_Trigoni_2018}, RIDI~\cite{yan2018ridi}, RoNIN~\cite{9196860}, ConvNet~\cite{cortes2018deep}, DeepVIP~\cite{zhou2022deepvip}, and LLIO~\cite{wang2022llio}. 
% A second line improves representation learning and generalization through stronger structural priors, including NILoc~\cite{herath2022neural}, IDOL~\cite{sun2021idol}, CTIN~\cite{rao2022ctin}, EqNIO~\cite{Jayanth25iclr}, and Tartan IMU~\cite{11094790}. 
% A third line combines learned inertial modeling with filtering or uncertainty-aware estimation, such as TLIO~\cite{liu2020tlio}, IMUNet~\cite{10480886}, AirIO~\cite{11045120}, AirIMU~\cite{qiu2024airimulearninguncertaintypropagation}, and IMO~\cite{10058169}. 
% These studies show that learning-based inertial tracking remains active in recent years, with progress in motion regression, geometric modeling, and hybrid estimation. 
% However, they are mainly developed for pedestrians, vehicles, or aerial robots, and are not tailored to the motion characteristics and deployment constraints of shared bikes.

On the other hand, many studies have established data-driven inertial motion models to mitigate the accumulation errors of SINS. 
\rev{Some studies focus on learning local motion variables directly from IMU sequences,} such as IONet~\cite{Chen_Lu_Markham_Trigoni_2018}, RIDI~\cite{yan2018ridi}, RoNIN~\cite{9196860}, ConvNet~\cite{cortes2018deep}, DeepVIP~\cite{zhou2022deepvip}, and LLIO~\cite{wang2022llio}. 
\rev{Others improve representation learning and generalization through stronger structural priors,} including NILoc~\cite{herath2022neural}, IDOL~\cite{sun2021idol}, CTIN~\cite{rao2022ctin}, EqNIO~\cite{Jayanth25iclr}, and Tartan IMU~\cite{11094790}. 
\rev{Others further combine learned inertial modeling with filtering or uncertainty-aware estimation,} such as TLIO~\cite{liu2020tlio}, IMUNet~\cite{10480886}, AirIO~\cite{11045120}, AirIMU~\cite{qiu2024airimulearninguncertaintypropagation}, and IMO~\cite{10058169}. 
\rev{These studies show that learning-based inertial tracking remains active in recent years, with progress in motion regression, geometric modeling, and hybrid estimation.}
\rev{However, they are mainly developed for pedestrians, vehicles, or aerial robots, and are not tailored to the motion characteristics and deployment constraints of shared bikes.}

In addition, several studies have explored multimodal fusion, employing DNNs to map multi-sensor data to poses. 
\cite{10330603} proposed an acoustic inertial measurement system that achieves indoor localization for drones by integrating acoustic features. 
\cite{lu2020heterogeneous} combined a convolutional autoencoder with a Temporal Convolutional Network (TCN) to process IMU noise and estimate latitude and longitude, which are used as observations for the KF.
Similarly, methods such as \cite{10806531, 10268655, huang2024visual} have demonstrated that integrating visual or LiDAR sensors can further improve localization performance, which is beyond the scope of this paper.

\subsection{Domain-Specific Knowledge}

Domain-specific knowledge in inertial navigation leverages prior information from application environments and motion patterns to enhance constraints and calibrate errors.   
Such approaches integrate pseudo-observations or structural constraints to improve the robustness and observability of localization in weak observation or high-noise scenarios.

% Pedestrian Dead Reckoning (PDR) employs the periodic patterns during pedestrian walking to track their motion, including step detection, heading and step length estimation, and position updates \cite{10716421}. Such methods analyze the waveform features of IMU signals and apply biomechanical models to extract gait parameters from inertial data \cite{10332483}. Zero-velocity update (ZUPT) employs methods such as local motion pattern modeling \cite{10137733} and learning-based phase detection \cite{goyal2011strap} to identify stationary phases during pedestrian walking, which are used to correct drifts in inertial navigation systems. These approaches utilize domain-specific knowledge in pedestrian location to extract discriminative features from inertial data, thus constraining error divergence. However, incorrect feature extraction compromises the accuracy of the estimate. Moreover, due to their reliance on motion-specific assumptions, these methods are limited to pedestrian tracking.

Pedestrian Dead Reckoning (PDR) exploits the periodic characteristics of human gait to track pedestrian positions through step detection, heading and step length estimation, and position updates~\cite{10716421}. 
Such methods typically integrate biomechanical models with IMU signal features to extract gait parameters~\cite{10332483}, and correct drift errors in inertial tracking with stationary phase detection (e.g., zero velocity update~\cite {10137733, 9991866}). 
Such methods rely on pedestrian-specific motion patterns, thus their localization accuracy depends on the quality of feature extraction, and their application scenarios are usually limited to pedestrian localization.

In addition, methods such as map matching~\cite{xia2023integrated} and non-holonomic constraints~\cite{zhang2024adaptive} are usually employed within navigation frameworks to constrain errors.
They are able to be regarded as reliable observations in specific scenarios.
However, they typically involve additional prior information, rendering them unsuitable for all situations.
In this paper, we identify a unique periodic pattern in riding scenarios and exploit this pattern as auxiliary information to improve the localization accuracy of shared bikes. 
To the best of our knowledge, this is the first work to discover this pattern.

\revtwo{Table~\ref{tab_related_comparison} compares the above inertial tracking methods in terms of carrier, model structure, error modelling and constraints. Existing methods mainly target pedestrians, vehicles, robots, UAVs, or sensor-level calibration, some of them estimating uncertainty through filtering or learning, or introduce constraints such as map matching to mitigate the divergence of errors. In contrast, we utilise a more robust model to learn motion updates and their uncertainties, combining these with wheel speeds extracted from riding behaviour, with a focus on tracking shared bikes.}

\begin{table}[t]
\centering
\caption{\revtwo{A Comparison of Prior Inertial Tracking Methods}}
\label{tab_related_comparison}
\scriptsize
\setlength{\tabcolsep}{2.5pt}
\renewcommand{\arraystretch}{1.0}
\begingroup
\ifShowChanges\color{red}\fi
\begin{tabular}{ccccc}
\toprule
\textbf{Reference} & 
\textbf{Carrier} & 
\textbf{Model} & 
\textbf{Error Modeling} & 
\textbf{Constraint} \\
\midrule
\makecell{\cite{Chen_Lu_Markham_Trigoni_2018, 9196860, yan2018ridi}, \\{} \cite{herath2022neural, 10480886}} &
Pedestrian &
\makecell{LSTM, ResNet,\\ TCN, SVM,\\ CNN, Transformer} &
- &
- \\
\midrule
\makecell{\cite{cortes2018deep, sun2021idol},\\{} \cite{11045120, 10058169, 10330603, lu2020heterogeneous, 10806531, 10268655, huang2024visual}} &
\makecell{Pedestrian,\\ UAV,\\ Vehicle,\\ Robot} &
\makecell{KF, CNN, LSTM,\\ GRU, TCN,\\ VIO, LIO} &
\makecell{Filter\\ covariance} &
\makecell{External\\ measurement} \\
\midrule
\makecell{\cite{wang2022llio, zhou2022deepvip},\\{} \cite{rao2022ctin, Jayanth25iclr, 11094790, liu2020tlio}} &
\makecell{Pedestrian,\\ Vehicle,\\ Robot} &
\makecell{MLP, LSTM,\\ ResNet,\\ Transformer,\\ CNN, GRU} &
\makecell{Learned\\ uncertainty} &
- \\
\midrule
\makecell{\cite{10137733, xia2023integrated},\\{} \cite{10716421, 10332483, 9991866, zhang2024adaptive}} &
\makecell{Pedestrian,\\ Vehicle} &
\makecell{SINS, PDR,\\ LSTM,\\ Kinematic model} &
- &
\makecell{Gait pattern,\\ zero velocity,\\ map matching,\\ NHC} \\
\midrule
\makecell{\cite{chen2018improving, li2022calib, nobre2019learning, brossard2020denoising, huang2022mems, qiu2024airimulearninguncertaintypropagation}} &
- &
\makecell{CNN, TCN,\\ GRU} &
\makecell{Bias/residual\\ compensation} &
\makecell{High-quality\\ reference} \\
\midrule
\textbf{Ours} &
\textbf{\makecell{Shared\\ bike}} &
\makecell{\textbf{MoE,}\\ \textbf{periodicity}\\ \textbf{detection}} &
\makecell{\textbf{Learned}\\ \textbf{uncertainty,}\\ \textbf{residual}\\ \textbf{compensation}} &
\makecell{\textbf{Pedal-wheel}\\ \textbf{coupling}} \\
\bottomrule
\end{tabular}
\endgroup
\end{table}

\section{Preliminary} \label{sec:preliminary}

In GNSS blocked environments, the current platform typically employs a Strapdown Inertial Navigation System (SINS) as the odometry solution for bike tracking.
Specifically, the system estimates motion variables such as orientation, velocity, and position through integration of inertial measurements from the bicycle.
The IMU provides the specific force $\hat{a} \in \mathbb{R}^{3}$ and angular rate $\hat{w} \in \mathbb{R}^{3}$, both expressed in the body frame.
\begin{equation}
\label{equa_1}
\left\{
\begin{aligned}
    \hat{a}&=a_{b}+R^{b}_{w} g_{w}+n_{a}, \\
    \hat{w}&=w^{w}_{b}+n_{w}.
\end{aligned}
\right.
\end{equation}
\noindent where $g_w$ denotes the gravitational acceleration in the world frame, $R^b_w$ denotes the rotation from the world frame to the body frame, $a_b$ is the linear acceleration in the body frame, and $w^w_b$ is the angular rate of the body frame relative to the world frame, expressed in the body frame. 
$n_a$ and $n_w$ represent measurement errors, which include both deterministic components (i.e., bias and scale factor errors) and random noise.

% 原图2 (中控加现实部署)
% \begin{figure}[t]
% \label{fig_bike_real_placement}
%     \begin{minipage}[b]{0.49\linewidth}
%     % \vspace{1pt}
%     \centerline{\includegraphics[width=\textwidth]{figs/bike.jpg}}
%     \centerline{(a) DiDi bike appearance}
%     \end{minipage}
%     \begin{minipage}[b]{0.49\linewidth}
%     % \vspace{1pt}
%     \centerline{\includegraphics[width=\textwidth]{figs/ideal_bike.jpg}}
%     \centerline{(b) Required parking region}
%     \end{minipage}
% \caption{The appearance of shared bikes from the DiDi ride-hailing platform and the required parking region along the road. The central controller is mounted on the bike with a pitch angle of fixed $22^{\circ}$.}
% \end{figure}

Traditional inertial navigation systems directly perform double integration of accelerations to obtain position, which results in unbounded drift due to the accumulation of measurement errors.
By contrast, SINS first updates attitude with accelerations and angular rates, providing accurate orientation for velocity and position integration.
The specific force measurements are transformed to the world frame and integrated once to obtain velocity, which is then further integrated to estimate position.
Although there are still two integrations, the attitude update significantly decreases the risk of divergence.

Intuitively, the tracking process consists of three stages: attitude update, velocity update, and position update.
Assuming ideal IMU measurements without errors as in Equation~\eqref{equa_1}, we have the following:
%%%
% \begin{equation}
% \label{equa_3}
% \left\{
%     \begin{aligned}
%         % \mathbf{R}_b^w(t+\Delta t) &= \mathbf{R}_b^w(t)\mathbf{R}_{b_{t+\Delta t}}^{b_t}\\
%         \mathbf{R}_b^w(t+\Delta t) &= \mathbf{R}_b^w(t)\rev{\mathbf{\Delta R}_t} \\
%         \mathbf{v}_w(t + \Delta t) &= \mathbf{v}_w(t) + \int_{t}^{t + \Delta t} \mathbf{a}_w(\tau) \, d\tau\\
%         % [\mathbf{R}_b^n(\tau) \cdot \mathbf{f}_b(\tau) - \mathbf{g}_n]
%         \mathbf{p}_w(t+\Delta t) &= \mathbf{p}_w(t)+\int_{t}^{t+\Delta t} \mathbf{v}_w(\tau) \, d\tau
%         % [\mathbf{v}_w(t)+\mathbf{a}_w(t)(\tau - t)]d\tau
%     \end{aligned}
% \right.
% \end{equation}
% \noindent where
% \begin{equation}
% \label{equa_4}
% \left\{
%     \begin{aligned}
%         \mathbf{a}_w(\tau) &= \mathbf{R}_b^w(\tau) \mathbf{a}_b(\tau) - \mathbf{g}_w,\\
%         \mathbf{v}_w(\tau) &= \mathbf{v}_w(t)+\int_t^{\tau}\mathbf{a}_w(s)\mathrm{d}s,
%     \end{aligned}
% \right.
% \end{equation}
%%%
\begin{equation}
\label{equa_3}
    \left\{
    \begin{aligned}
    \mathbf{R}_b^w(t+\Delta t)&=\mathbf{R}_b^w(t)\rev{\mathbf{\Delta R}_t},\\
    \mathbf{v}_w(t+\Delta t)&=\mathbf{v}_w(t)+\int_t^{t+\Delta t}\mathbf{a}_w(\tau)\,\mathrm{d}\tau,\\
    \mathbf{p}_w(t+\Delta t)&=\mathbf{p}_w(t)+\int_t^{t+\Delta t}\mathbf{v}_w(\tau)\,\mathrm{d}\tau,
    \end{aligned}
    \right.
\end{equation}
\noindent where
\begin{equation}
\label{equa_4}
    \left\{
    \begin{aligned}
    \mathbf{a}_w(\tau)&=\mathbf{R}_b^w(\tau)\mathbf{a}_b(\tau)-\mathbf{g}_w,\\
    \mathbf{v}_w(\tau)&=\mathbf{v}_w(t)+\int_t^\tau \mathbf{a}_w(s)\,\mathrm{d}s.
    \end{aligned}
    \right.
\end{equation}
\noindent where $\mathbf{R}_b^w(t)\in SO(3)$ denotes the rotation from the body frame to the world frame 
\rev{at time $t$, $\mathbf{\Delta R}_t\in SO(3)$ denotes the incremental rotation from gyroscope measurements over the interval $[t,t+\Delta t]$,}
and $\mathbf{g}_w \in \mathbb{R}^3$ is the gravity vector in the world frame.
%
% \noindent where $\mathbf{R}_b^w(t) \in SO(3)$ denotes the rotation from the body frame to the world frame, and $\mathbf{g}_w \in \mathbb{R}^3$ is the gravity vector in the world frame.
% In practice, inertial readings are typically denoised with low-pass filters, and integrated with available GNSS signals by techniques such as the Kalman Filter (KF) to calibrate localization states.

\rev{Although SINS exhibits favourable interpretability, it remains dependent on the continuous propagation of attitude, velocity, and position. 
For generic low-cost IMUs, sensor noise, bias, scale factor errors, and mounting errors accumulate during this open-loop process, leading to rapid drift. 
In particular, even small attitude errors can distort gravity compensation and further amplify velocity and position errors.}

\rev{To alleviate error accumulation, recent learning-based methods, such as IONet~\cite{Chen_Lu_Markham_Trigoni_2018}, RoNIN~\cite{9196860}, and LLIO~\cite{wang2022llio}, reformulate inertial tracking by replacing continuous integration-based state propagation with the estimation of short-term motion updates, such as velocity, displacement, or heading increments, from fixed-window IMU sequences for incremental trajectory reconstruction. 
In this sense, these methods utilise learning-based alternatives to replace the integration-based motion estimation process, rather than correction modules for certain states within a conventional inertial navigation framework. 
Inspired by them, we regard the tracking of shared bikes in GNSS-blocked environments as a short-term inertial motion estimation task, and innovatively extract opportunistic speed observations to calibrate long-term errors.}

\section{Design Overview}\label{sec:overview}

% In large-scale shared bike platforms, hardware configurations must balance manufacturing cost, maintenance efficiency, and lifecycle management. 
% Constrained by such industrial factors, shared bikes are typically equipped only with an IMU, a GNSS positioning module, and a mobile communication module. 
% Figure \ref{fig_bike_real_placement}(a) illustrates the overall appearance of a shared bike from DiDi in China, along with the mounting angle of its central controller, which integrates above sensors. 
% The communication module supports remote connectivity, including uploading data and receiving commands.   
% Figure \ref{fig_bike_real_placement}(b) shows the actual deployment of shared bikes in real-world scenarios.
% When bikes enter GNSS blocked environments, satellite observations become unreliable, thus inertial measurements are the primary sensing modality of motion estimation. 
% However, IMU noises accumulate rapidly during integration, causing unbounded errors for bike tracking. 

In large-scale shared bike platforms, hardware configurations must balance manufacturing cost, maintenance efficiency, and lifecycle management. 
\rev{Under such constraints, shared bikes are typically equipped with a central controller integrating an IMU, a GNSS positioning module, and a mobile communications module.}
The communication module supports remote connectivity for data uploading and command reception. 
In real-world deployment, when bikes enter GNSS-blocked environments, satellite observations become unreliable, making inertial measurements the primary modality for motion estimation. 
However, IMU noise accumulates rapidly during integration, leading to unbounded tracking errors.

\begin{figure}[t]
\label{fig_overview}
    \centering
    \begin{minipage}{0.9\linewidth}
    % \vspace{3pt}
    \centerline{\includegraphics[width=\textwidth]{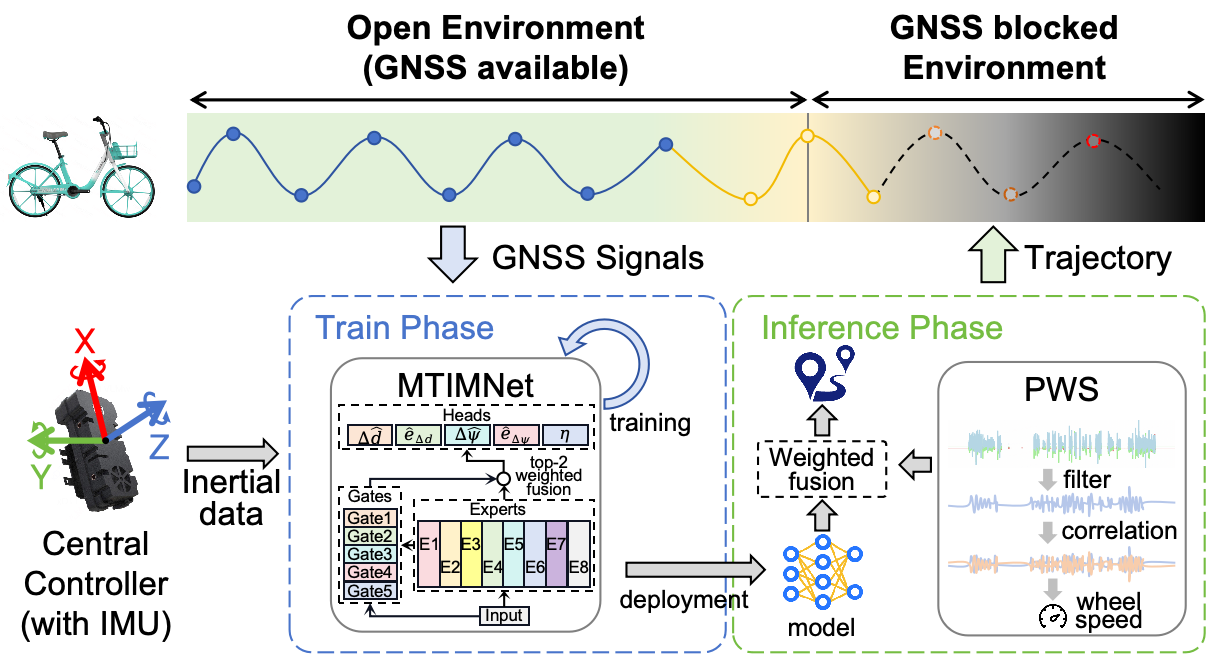}}
    \end{minipage}
    \caption{Overview of the proposed inertial tracking framework for shared bikes. It consists of a Multi-Task Inertial Motion Network (MTIMNet) and a pseudo wheel speed (PWS) module. MTIMNet is trained with GNSS signals in open environments and used for GNSS-free inference, while PWS provides velocity calibration during long-term tracking.}
\end{figure}

% In this paper, we propose an inertial tracking framework for shared bikes that only relies on the bike's IMU readings in GNSS blocked environments.
% Figure \ref{fig_overview} depicts our framework, which consists of a multi-task inertial motion network (MTIMNet, Section \ref{sec:mtimnet}) and a pseudo wheel speed estimation method for calibration (PWS, Section \ref{sec:wheel}).
% Specially, the first network is trained on riding data with GNSS as the ground truth in open areas, and infers the bike trajectory from realtime inertial sequences in GNSS blocked environments. The second method runs directly on accelerations without training, 
% as it extracts bike's speed from the rider's periodic pedalling behavior.
% In a word, GNSS signals are only used during training, but not engaged in location inference.

In this paper, we propose an \rev{IMU-only inertial tracking framework for shared bikes} in GNSS-blocked environments. 
As shown in Figure~\ref{fig_overview}, it consists of a multi-task inertial motion network (MTIMNet, Section~\ref{sec:mtimnet}) and a pseudo wheel speed estimation method (PWS, Section~\ref{sec:wheel}). 
The MTIMNet is trained on riding data with GNSS as ground truth in open areas, \rev{learning short-term motion variables from IMU sequences for incremental trajectory reconstruction, replacing the integration process of conventional SINS.} 
The PWS method derives auxiliary speed observations directly from periodic pedalling patterns in accelerations without training. 
\rev{Thus, GNSS is used only for training supervision, whereas inference in blocked environments relies entirely on inertial measurements.}

In practice, the embedded devices on shared bikes have limited computational resources and lack mature deep learning frameworks, making it difficult to execute complex models directly on the bike. 
Therefore, inertial readings are uploaded to remote servers at fixed intervals (e.g., every second) for model inference, \rev{while the pseudo wheel speed module runs locally on the bike. 
Once valid pseudo wheel speed estimates are obtained, they are fused with the output of the model using a variance-weighted method. 
This strategy enables effective deployment with low computational cost on the device.}

% old
% In our industrial deployment, the embedded devices on shared bikes are constrained by limited computational resources and lack of mature deep learning frameworks, making it difficult to execute complex models directly on the bike side.
% Practically, large-scale shared bikes collect inertial readings continuously and upload data to remote servers at fixed intervals (e.g., every second) for model inference.
% This strategy ensures an effective model deployment while reducing the computational load on the device.

\section{MTIMNet Design}\label{sec:mtimnet}

For shared bikes, MTIMNet explores an inertial sequence learning method that leverages realtime inertial readings to infer the bike's location, with corresponding confidence assessment on such estimates.

\subsection{Problem formulation}\label{problem_desc}

\rev{Based on Section~\ref{sec:preliminary}, we replace the integration-based motion estimation in conventional SINS with a learning-based estimation of short-term motion variables from IMU sequences, and progressively reconstruct the bicycle trajectory in GNSS-blocked environments. 
Since the task mainly involves planar motion, trajectory updates are primarily determined by displacement and heading increments, as follows:}

\begin{equation}
    \label{equa_position}
    \left\{
    \begin{aligned}
    p^x_{t+\Delta t} &= p^x_t + \Delta d_t \sin(\psi_{t+\Delta t}),\\
    p^y_{t+\Delta t} &= p^y_t + \Delta d_t \cos(\psi_{t+\Delta t}),\\
    \psi_{t+\Delta t} &= \psi_t + \Delta \psi_t.
    \end{aligned}
    \right.
\end{equation}
\noindent where $(p^x, p^y)$ denotes the position in the plane, $\Delta d_t$ and $\Delta \psi_t$ denote the displacement and heading increment over the interval $[t, t+\Delta t]$.
% with due north as 0 degrees $(0^{\circ})$. 

In addition, IMU measurement errors are typically composed of deterministic components, including bias, scale factors, and random components such as random walk noise~\cite{chen2024deep}.
Existing work always approximates these errors with a Gaussian distribution.
However, in practical riding scenarios, complex environments and low-cost IMUs tend to introduce additional systematic biases, such as mounting offsets or prediction bias that deviate from the Gaussian assumption. 
To better capture these uncertain effects, we further decompose the error sources as:
\begin{equation}
\label{equa_6}
    \varepsilon = e + \eta, \quad \eta \sim \mathcal{N}(0, \sigma^2)
\end{equation}
\noindent where $\varepsilon$ denotes the overall error, \rev{$e$} represents the learnable bias, and $\eta$ is an unpredictable noise term that approximately follows a Gaussian distribution. 
Therefore, \rev{unlike existing approaches, MTIMNet leverages the residual to correct deterministic bias and captures random uncertainty by variance. The outputs for displacement and heading increments are:}
\begin{equation}
\label{equa_7}
    y = \hat{y} + e + \eta
\end{equation}
\noindent where $y$ denotes the displacement or heading increment, $\hat{y}$ denotes the initial estimate, $e$ represents the residual estimate, and $\eta$ denotes the uncertainty estimate.
\rev{The residual term models deterministic bias, while the uncertainty term captures the remaining random error. 
They correspond to different error components and play different roles in the task.}

\subsection{Model structure}

% old
% Existing approaches such as IONet \cite{Chen_Lu_Markham_Trigoni_2018}, RoNIN \cite{9196860}, and LLIO \cite{wang2022llio} have leveraged DNNs to directly learn local motion variables like displacement and velocity, thus mitigating error accumulation from double integration.
% However, they typically employ one module (e.g., a multilayer perceptron) to estimate multiple variables simultaneously, limiting their ability to model correlations between variables.
% In addition, \cite{chen2019deep, wang2022llio, rao2022ctin} assume that model errors follow a Gaussian distribution estimate model uncertainty, whereas this assumption is overly idealized, making it difficult to compensate errors with complex noises and severe outliers in low-cost IMU.

\begin{figure}[t]
    \begin{minipage}[b]{0.335\linewidth}
    % \vspace{1pt}
    \centerline{\includegraphics[width=\textwidth]{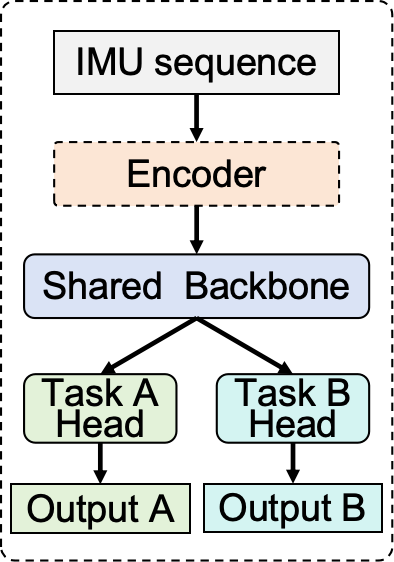}}
    \centerline{(a) Existing Model}
    \end{minipage}
    \begin{minipage}[b]{0.655\linewidth}
    % \vspace{1pt}
    \centerline{\includegraphics[width=\textwidth]{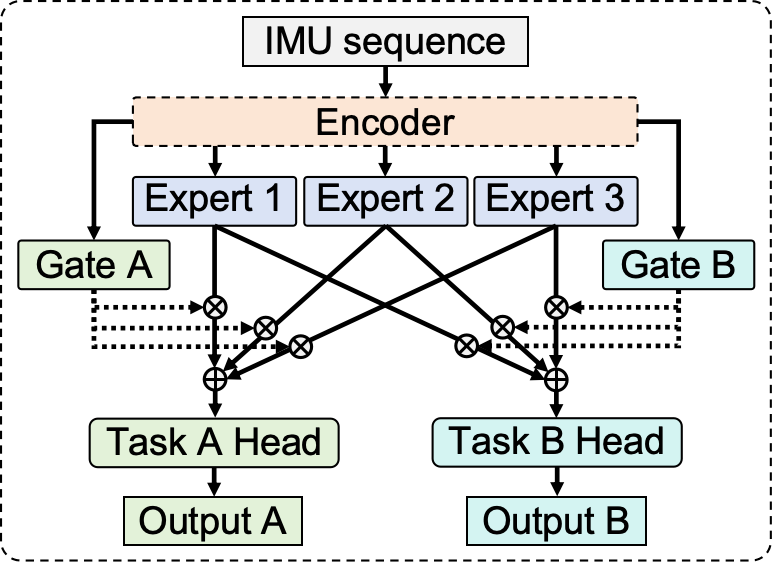}}
    \centerline{(b) MMoE-style Model}
    \end{minipage}
\caption{\rev{Model architecture comparison between the conventional multi-head shared backbone network and the MMoE-style network.}}
\label{fig_model_comparison}
\end{figure}

Existing approaches such as IONet~\cite{Chen_Lu_Markham_Trigoni_2018}, RoNIN~\cite{9196860}, and LLIO~\cite{wang2022llio} leverage DNNs to directly estimate short-term motion updates from IMU sequences, alleviating error accumulation from conventional inertial integration.
\rev{However, they typically adopt a shared backbone network with similar multi-head outputs, where multiple motion-related variables are still learned from shared latent features, as shown in Figure~\ref{fig_model_comparison}(a).}
This coarse task decomposition may introduce interference between related tasks.
Moreover, although methods such as \cite{chen2019deep, wang2022llio, rao2022ctin} further estimate uncertainty together with motion updates, \rev{they do not explicitly distinguish deterministic bias from random uncertainty, which limits their ability to model the complex error characteristics of low-cost IMU data.}

Recently, Multi-gate Mixture-of-Experts (MMoE)~\cite{10.1145/3219819.3220007} has shown that different tasks in multi-task learning usually benefit from both shared features and task-specific features, rather than relying on fully shared or fully separate modeling alone. 
Inspired by this observation, we construct the \textbf{M}ulti-\textbf{T}ask \textbf{I}nertial \textbf{M}otion \textbf{Net}work (\textbf{MTIMNet}), which adopts the MMoE-style routing mechanism shown in Figure~\ref{fig_model_comparison}(b). 
\rev{Different from conventional multi-head architectures with a fixed shared backbone, MTIMNet uses shared experts to learn diverse motion features and task-specific gates to adaptively fuse them for each task, enabling more effective cross-task feature sharing while reducing interference between tasks.}
% old
% Recently, Multi-gate Mixture-of-Experts (MMoE) \cite{10.1145/3219819.3220007} indicates that task relationships in multi-task learning are typically neither completely independent nor fully aligned, but partially shared and partially specific. 
% Inspired by this, we construct a \textbf{M}ulti-\textbf{T}ask \textbf{I}nertial \textbf{M}otion \textbf{Net}work (\textbf{MTIMNet}) with a mixture-of-experts (MoE) mechanism. 
% \rev{Unlike conventional multi-head architectures with a fixed shared backbone, MTIMNet adopts a mixture-of-experts structure for more flexible task relationship modeling. 
% Shared experts learn diverse motion features, and task-specific gates adaptively fuse them for each task, enabling more effective cross-task feature sharing while reducing interference between tasks.}

\begin{figure}[t]
\label{fig_model}
    \centering
    \begin{minipage}{0.9\linewidth}
    % \vspace{3pt}
    \centerline{\includegraphics[width=\textwidth]{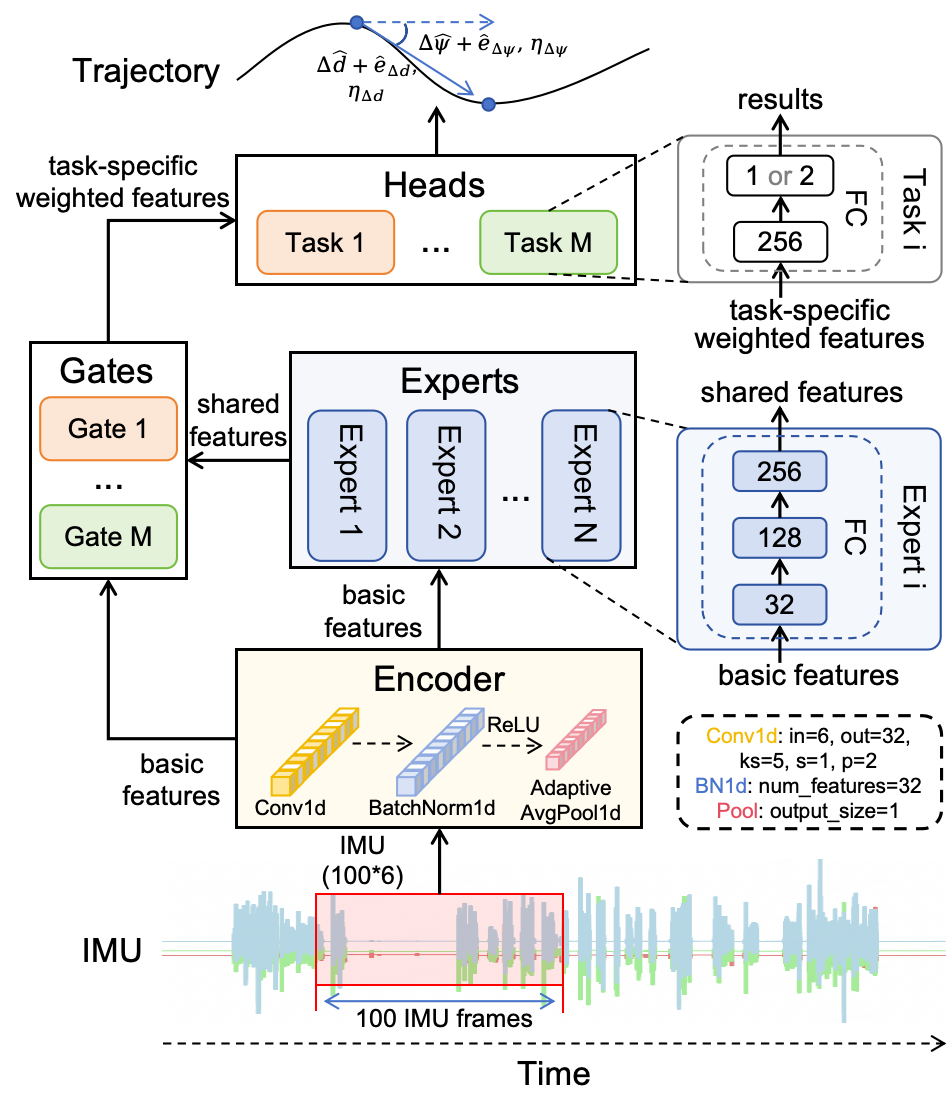}}
    \end{minipage}
    \caption{Overall architecture of the MTIMNet. An encoder extracts features from IMU windows, \rev{which are processed by shared experts and task-specific gates for multi-task regression.} In this paper, $M=5$ represents five tasks, including displacement $\hat{\Delta d}$ and its residual $\hat e_{\Delta d}$, heading increment $\hat{\Delta \psi}$ and its residual $\hat e_{\Delta \psi}$, and uncertainty $\eta$. \rev{$N=8$ represents the number of experts.}}
\end{figure}

The architecture of MTIMNet is illustrated in Figure~\ref{fig_model}. 
It consists of an inertial encoder module, multiple expert modules, multiple gating modules, and multiple task-specific heads.
\rev{The encoder first maps each IMU window into latent features, which are then fed to all shared experts and task-specific gates. 
The experts generate candidate shared features, while the gates predict task-specific routing weights over these expert outputs. 
For each task, the features from the top two experts in terms of weighting will be weighted and fused, and then passed to the corresponding task head for prediction.}

The \textbf{inertial encoder module} employs a stacked architecture consisting of a convolutional layer, a batch normalization layer, and an average pooling layer.
The convolutional layer extracts basic local features from the inertial sequence, while the normalization and pooling layers mitigate distribution bias and reduce redundant information, respectively.

The \textbf{expert module} consists of multiple fully connected layers. We utilize eight experts to learn task-independent deep shared representations from basic features, enabling the model to capture latent patterns across different motion modes.

% old
% The \textbf{gating module} contains a fully connected layer, a softmax layer, and a dropout layer, as shown in Figure \ref{fig_routing}. It dynamically calculates the weights of each expert and performs a weighted fuse of the features from the top-2 experts with the highest weights.
% Such structure prevents over-dominance by certain experts while enhances the model's robustness.

\begin{figure}[b]
\label{fig_routing}
    \centering
    \begin{minipage}{0.9\linewidth}
    % \vspace{3pt}
    \centerline{\includegraphics[width=\textwidth]{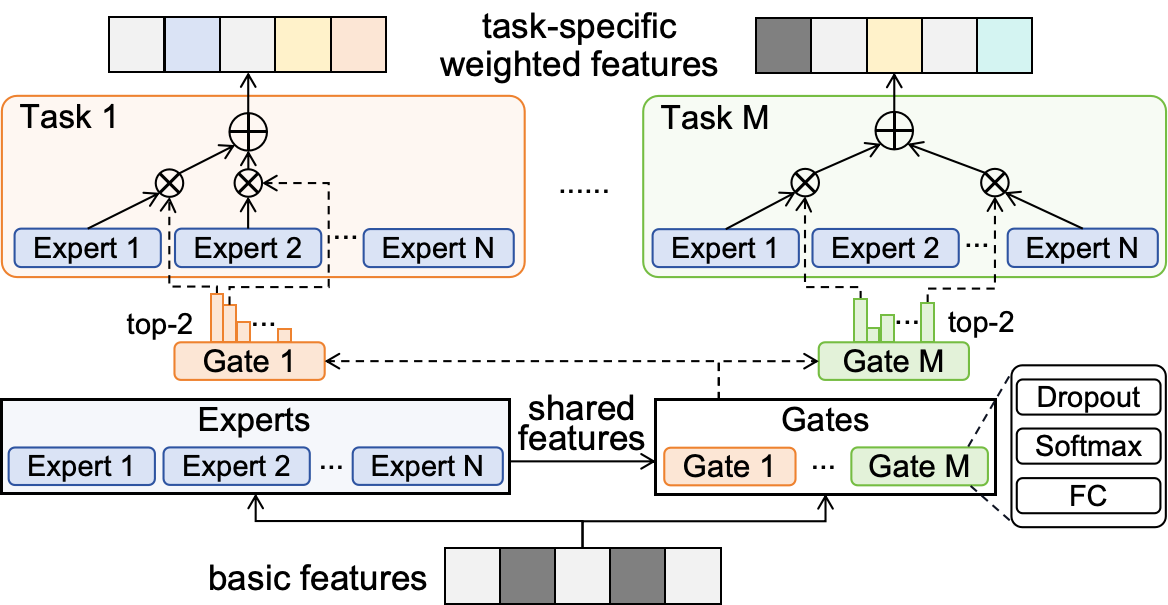}}
    \end{minipage}
    \caption{\rev{Illustration of task-specific gating and top-2 expert fusion in MTIMNet. For each task, a task-specific gate predicts routing weights over all experts and fuses the outputs of top-2 experts with the highest weights.}}
\end{figure}

The \textbf{gating module} contains a fully connected layer, a softmax layer, and a dropout layer, as shown in Figure~\ref{fig_routing}.
\rev{It dynamically estimates the weight of each expert and obtains task-specific weighted features by fusing the features of the two experts with the highest weights.
This design balances expert specialization and limited collaboration by avoiding both overly sparse single-expert routing and redundant multi-expert fusion.}

The \textbf{task-specific head} exploits two fully connected layers to estimate the target from weighted features.
Specifically, here are five tasks: 
\rev{
\begin{itemize}
    \item Task 1: displacement $\Delta \hat{d}$,
    \item Task 2: heading increment $\Delta \hat{\psi}$,
    \item Task 3: displacement residual $\hat{e}_{\Delta d}$,
    \item Task 4: heading increment residual $\hat{e}_{\Delta \psi}$,
    \item Task 5: uncertainty term $\mathcal{\eta}$, consisting of $\mathcal{\eta}_{\Delta d}$ and $\mathcal{\eta}_{\Delta \psi}$.
\end{itemize}
}

% 原图注: Overview of the gating module. Task-specific gates estimate the weights of different experts and fuse the features of the top two experts with the highest weights.

\rev{Rather than deriving all task features from a single shared backbone network, MTIMNet utilises multiple experts and task-specific routing mechanisms to capture both shared and task-specific motion patterns.}
In practice, MTIMNet serves as a function \rev{$f$} that maps inertial readings to motion variables within a window of $\Delta t$:
\begin{equation}
\label{equa_8}
    f(\mathbf{X}_{t - \Delta t: t}) \rightarrow \left( \Delta \hat{d}, \Delta \hat{\psi}, \hat{e}_{\Delta d}, \hat{e}_{\Delta \psi}, \mathcal{\eta}_{\Delta d, \Delta \psi} \right)
\end{equation}
\noindent where $X$ denotes IMU data, and $\Delta t$ represents 100 frames (1-second).

\subsection{Loss function}

We design a composite loss function to guide model training, which encourages two properties:
(i) confidence-aware motion estimation, and (ii) consistency of the residual estimates. 
For each target $y \in \{\Delta d, \Delta \psi\}$ of a shared bike, the loss consists of two terms:
\begin{equation}
\label{equa_9}
    \begin{aligned}
        \mathcal{L} = \underbrace{\frac{1}{2 \exp(\eta)} \lVert y - \hat{y} \rVert^2 + \frac{1}{2} \eta 
        \vphantom{\frac{1}{2\exp(\eta)}}  % 高度参照
        }_{\text{Negative log-likelihood (NLL)}} 
        + \underbrace{L_{SmoothL1}(\hat{e}, (y - \hat{y}))
        \vphantom{\frac{1}{2\exp(\eta)}}  % 高度参照
        }_{\text{Residual estimation loss}}
    \end{aligned}
\end{equation}
\noindent where $\hat{y}$ and $\eta$ are the mean value and log-variance of model prediction, respectively. As regressing the log-variance is more stable than directly predicting the variance~\cite{10.5555/3295222.3295309}, the first loss term adopts a negative log-likelihood formulation under the log-variance parameterization to represent model uncertainty. 
The second term uses Smooth L1 Loss $L_{SmoothL1}$ to minimize the difference between the predicted residual $\hat{e}$ and its measurement ($y-\hat{y}$) for self-correction.

\rev{This formulation differs from existing multi-head regression. 
The residual branch models deterministic bias, while the uncertainty branch estimates the confidence of the remaining error. 
In this way, the MTIMNet is trained to capture deterministic and random errors separately within a unified framework.}
In sum, the total loss is defined as:
\begin{equation}
\label{equa_10}
    L_{total} = \sum_{y \in \{ \Delta d, \Delta \psi \}} \mathcal{L}_y
\end{equation}
%The predicted state, uncertainty and residual play different but complementary roles. 

To track the shard bike, we summarize the model estimate with corresponding residual to produce the final location.

\section{Pseudo Wheel Speed Estimation}\label{sec:wheel}

% Although MTIMNet infers bike trajectories from inertial readings, long-term error accumulation inevitably leads to unbounded location errors.
% It is essential to calibrate model estimation via reliable observations, e.g., the bike velocity.

Although learning-based inertial estimation reduces the error accumulation of conventional inertial navigation, long-term inference still inevitably suffers from drift. 
Therefore, additional observations are needed to further constrain error growth during extended tracking, e.g., the bike velocity.

However, obtaining such observations from shared bikes always remains challenging.
Sensors such as wheel odometers are difficult in large-scale deployment due to cost constraints, whereas error calibration methods (e.g., zero velocity updates~\cite{10137733}) exhibit dependence on specific scenarios, thus limiting their cross-scene availability.
To fill this gap, 
\rev{we design a pseudo wheel speed (PWS) estimation method that detects periodic pedalling behavior from inertial data and converts it into velocity observations for calibration, thereby constraining long-term drifts.}
% we propose a pseudo wheel speed (PWS) estimation method, which derives bicycle speed by detecting the rider’s periodic pedalling pattern from inertial data.

%\begin{figure}[t]
%\label{fig_period_raw_filtered_acc_data}
    %\begin{minipage}[b]{0.49\linewidth}
    % \vspace{1pt}
    %\centerline{\includegraphics[width=\textwidth]{figs/period_bike.jpg}}
    %\centerline{(a) Pedaling forces area}
    %\end{minipage}
    %\begin{minipage}[b]{0.49\linewidth}
    % \vspace{1pt}
    %\centerline{\includegraphics[width=\textwidth]{figs/raw_filtered_acc_data.jpg}}
    %\centerline{(b) Acceleration processing}
    %\end{minipage}
%\caption{An overview of the force-generating area during riding, as well as a periodic representation of the acceleration data.}
%\end{figure}

\begin{figure}[t]
    \centering
    \begin{minipage}{0.75\linewidth}
    % \vspace{3pt}
    \centerline{\includegraphics[width=\textwidth]{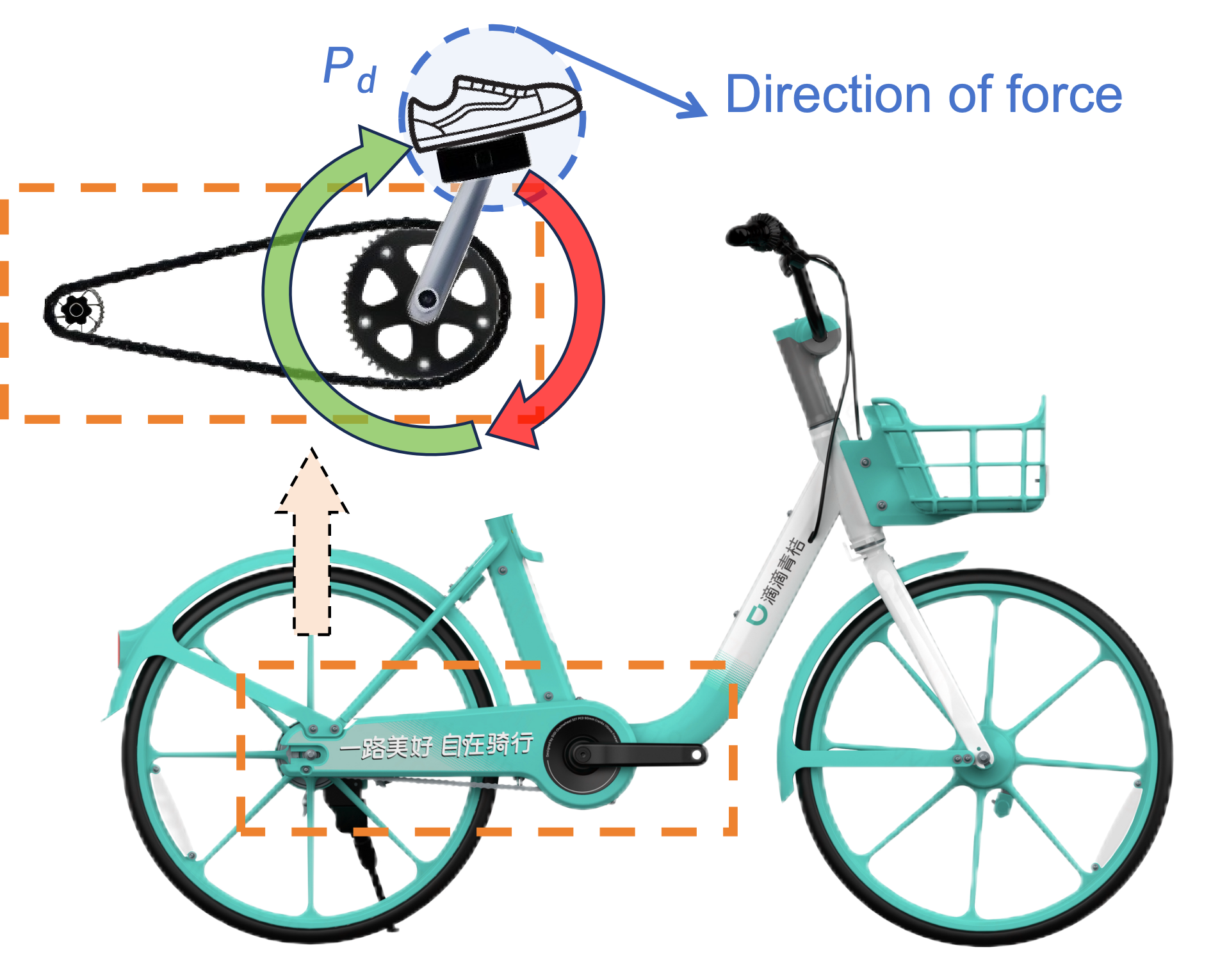}}
    \end{minipage}
    \caption{Demonstration of the pedal action and one-way clutch mechanism. People exert more force in the red region than in the green region.}
    \label{fig_bike_force}
\end{figure}

\subsection{Observation: Periodic pedalling during riding}

\rev{Intuitively, we observe that the force exerted during riding is not uniform, but varies systematically over the pedalling period, as similarly observed in \cite{OBRYAN201411} on muscle coordination and power output during riding.}
As shown in Figure~\ref{fig_bike_force}, a ride generally exerts larger force when the pedal passes through the front region (marked as red), and less force in left region (marked as green). $P_d$ denotes the typical position where a rider begins to exert force on the pedal.

Specifically, a one-way clutch mechanism exists between the pedal and the rear wheel of the shared bike, as indicated by the orange box in Figure~\ref{fig_bike_force}. 
\revtwo{This means that each forward pedal rotation drives the rear wheel through a fixed number of revolutions, corresponding to an approximately constant transmission ratio $K$. 
Together with the rear-wheel radius $r$, these parameters are treated as platform/model-level mechanical constants\footnote{\revtwo{In practice, $K$ and $r$ are platform/model-level mechanical parameters. Since each shared-bike platform usually contains only a few bike models with fixed transmission ratios and wheel sizes, these parameters can be obtained from specifications or calibrated in advance.}}.}
In practice, mechanical resistance and bike condition may affect the conversion accuracy, but not the existence of the pedal-induced periodic pattern itself.

% raw
% Thus, we calculate the rotation speed of a bike's rear wheel, i.e. the bike's velocity, from the number of pedalling cycles.
% Assuming that the radius of the rear wheel is $r$, and the time for a complete pedalling cycle is $T$, the average speed $v$ of the bike during this cycle is expressed as:
Thus, we calculate the rotation speed of the rear wheel, i.e., the bike velocity, from the number of pedalling cycles.
Assuming that the time for a complete pedalling cycle is $T$, the average speed $v$ of the bike during this cycle is:
\begin{equation}
\label{equa_period_v}
    v = \frac{2\pi rK}{T}
\end{equation}

% \rev{It should be noted that the proposed pseudo wheel speed is only activated within intervals where effective periodic pedalling is detected. Therefore, it is regarded more as an auxiliary observation rather than a continuously available source of velocity.}

\subsection{Periodic detection}

Calculating the pedalling cycle $T$ is not easy, especially when tracking bikes in real time. Thus, we explore an autocorrelation-based period detection method to extract the average pedalling interval $\bar{T}$ from forwarding accelerations.

\begin{figure}[t]
\label{fig_period_peocess_flow}
    \centering
    \begin{minipage}{1.0\linewidth}
        \centerline{\includegraphics[width=\textwidth]{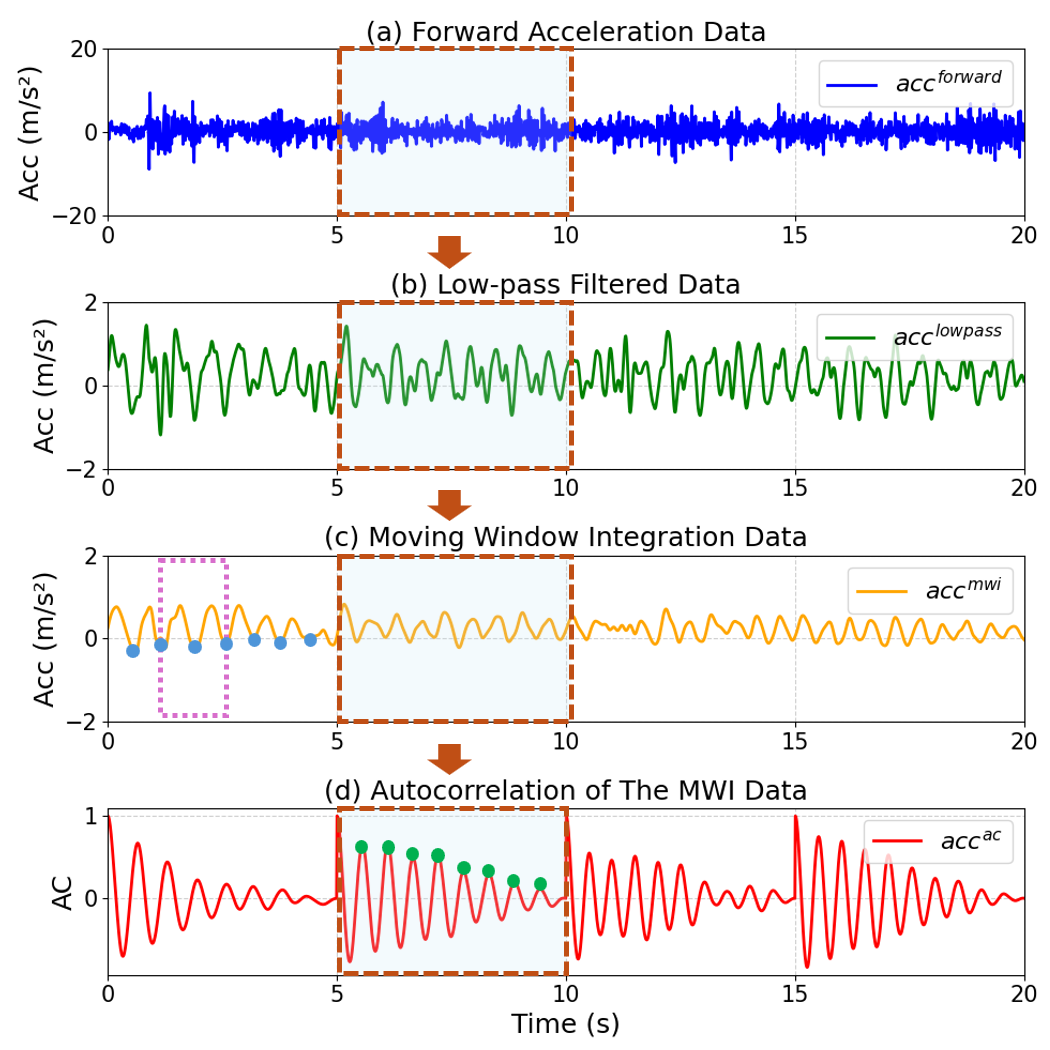}}
    \end{minipage}
    \caption{Calculation process of pedalling cycles. From top to bottom: forward accelerations, low-pass filtered acceleration data, Moving Window Integration(MWI) sequence, and the autocorrelation result of the MWI sequence. 
    We adopt a 5 seconds forward acceleration window to calculate the rider's average pedalling cycle (brown box).}
\end{figure}

Specifically, this period detection method mainly consists of four steps: forward acceleration calculation, low-pass filtering, moving window integration, and autocorrelation estimation, as illustrated in Figure~\ref{fig_period_peocess_flow}.

First, we calculate the forward acceleration $acc^{forward}$ (Figure~\ref{fig_period_peocess_flow}(a)) from the x-axis and z-axis of raw accelerations to eliminate interference from irrelevant signals, i.e.:
\begin{equation}
\label{equa_period_acc_forward}
    acc^{forward}=acc^z \cdot \cos{\theta}-acc^x \cdot \sin{\theta}
\end{equation}
\noindent where $\theta$ denotes the IMU mounting angle. 

Second, we denoise $acc^{forward}$ with a low-pass filter~\cite{butterworth1930theory}, yielding $acc^{lowpass}$ (Figure~\ref{fig_period_peocess_flow}(b)).

Third, we further enhance the periodic pattern of $acc^{lowpass}$ by Moving Window Integration (MWI)~\cite{ying2007automatic}, resulting in $acc^{mwi}$.
\begin{equation}
\label{equa_period_filter}
    acc^{mwi}_{i} = \frac{1}{N} \sum_{j=-k}^{k-1} acc_{i+j}
\end{equation}
\noindent where $k$ is the half length of time window, and $N=2k$ represents the total number of samples within this window. In our implementation, we adopt $k=15$.

Figure~\ref{fig_period_peocess_flow}(c) briefly demonstrates the pedalling behavior of $acc^{mwi}$, where the purple box denotes a complete pedalling cycle.  
Each acceleration valley corresponds to the moment when a rider begins exerting force on the pedal, and the intervals between two acceleration valleys approximately indicate a half cycle of pedal rotation.

Finally, we compute the autocorrelation~\cite{box2015time} of $acc^{mwi}$ to determine its periodicity, yielding the sequence $acc^{ac}$ (Figure~\ref{fig_period_peocess_flow}(d)).

The interval between peaks in $acc^{ac}$ corresponds to the period of $acc^{mwi}$. We calculate the time difference between adjacent peak moments to construct $t_{diff}$. If $\frac{max(t_{diff})}{min(t_{diff})}<1.4$, the current data is considered to exhibit periodicity. 
Specifically, we compute the average pedalling cycle time as $\bar{T}_{acc^{mwi}} = mean(t_{diff})$, thus leading to $\bar{T}=2 \cdot \bar{T}_{acc^{mwi}}$.

\subsection{Anomaly detection}
In practice, we observe that when a rider pedals continuously at high speeds (e.g., exceeding $4.5 m/s$), the pedals may fail to simultaneously rotate with the rear wheel.

\begin{figure}[t]
\label{fig_period_fast_mwi_ac}
    \begin{minipage}[b]{0.49\linewidth}
    % \vspace{1pt}
    \centerline{\includegraphics[width=\textwidth]{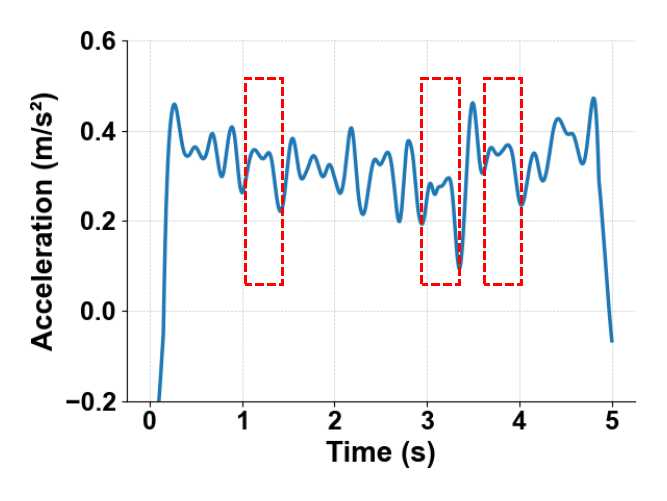}}
    \vspace{-5pt}
    \centerline{(a) MWI sequence}
    \end{minipage}
    \begin{minipage}[b]{0.49\linewidth}
    % \vspace{1pt}
    \centerline{\includegraphics[width=\textwidth]{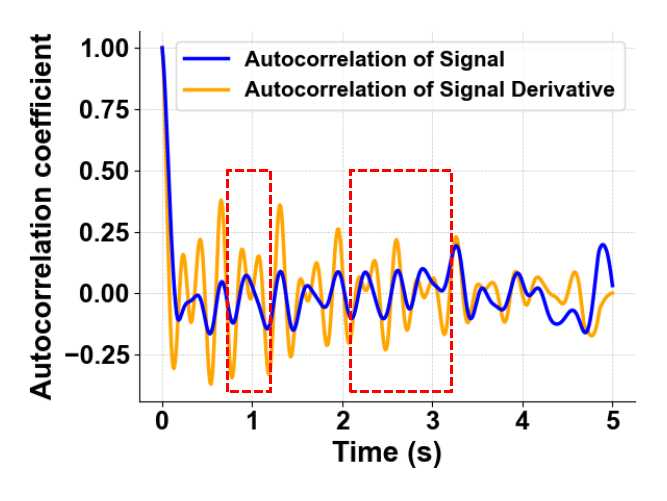}}
    \vspace{-5pt}
    \centerline{(b) Autocorrelation sequence}
    \end{minipage}
\caption{The MWI and autocorrelation sequences of forward accelerations in the fast riding scenario.}
\end{figure}

Normally, the data marked by the red box in Figure~\ref{fig_period_fast_mwi_ac}(a) should contain only one peak. However, the above anomaly leads to multiple peaks here.
To filter such anomalies, we first calculate the derivative of $acc^{mwi}$, denoted as $\dot{acc}^{mwi}$, and then further compute autocorrelation values of $\dot{acc}^{mwi}$, denoted as $\dot{acc}^{ac}$.
In abnormal scenarios, $acc^{ac}$ and $\dot{acc}^{ac}$ exhibit significant deviations, as illustrated in Figure~\ref{fig_period_fast_mwi_ac}(b).
We identify these deviations with the following two methods.

\begin{itemize}
    \item Pearson Correlation Coefficient (PCC): We exploit the Pearson correlation coefficient $\rho$ to measure the similarity between the two autocorrelation sequences. The two signals are considered different if $\rho<0.6$.
    \item Comparison of Peak Counts (CPC): We denote the number of peaks within a data window by $peaks(\cdot)$. The two signals are considered different if $\frac{peaks(\dot{acc}^{ac})}{peaks(acc^{ac})}>1.2$.
\end{itemize}

Once a valid average pedalling period $T=\bar{T}$ is extracted from anomaly-free data, the current wheel speed can be calculated with Equation~\eqref{equa_period_v}.

\subsection{Fusion with MTIMNet estimation}
\rev{In practice, several riding conditions may violate the steady-pedalling assumption, including reverse pedalling, coasting, downhill motion without pedalling, and other sections where speed varies. 
Therefore, pseudo wheel speed is estimated only when valid periodic pedalling patterns are detected.}

\rev{Once a valid pseudo wheel speed is detected\footnote{Since the proposed pseudo wheel speed is available only when effective periodic pedalling is detected, it is regarded as an auxiliary observation rather than a continuously available speed measurement.}, 
it is converted into a displacement observation over the corresponding interval and fused with the MTIMNet displacement estimate.
Let $\hat{d}_m$ and $\sigma_m^2$ be the displacement and variance predicted by MTIMNet, and let $\hat{d}_p$ and $\sigma_p^2$ be those derived from pseudo wheel speed, where $\sigma_p^2$ is derived from the standard deviation of the speed error for different riding speed levels (i.e. slow, moderate and fast) as detailed in Section~\ref{sec_eval_pws}.
The fused displacement is given by:}

\begin{equation}
\rev{\hat{d}_f=\frac{\hat{d}_m/\sigma_m^2+\hat{d}_p/\sigma_p^2}{1/\sigma_m^2+1/\sigma_p^2}.}
\end{equation}

\rev{This inverse-variance weighted fusion assigns a larger weight to the more reliable estimate and helps reduce long-term drift.}
\revtwo{When no valid PWS observations are available, the system skips the fusion and directly utilises the MTIMNet estimates. 
Consequently, PWS serves as an opportunistic calibration source rather than a continuously available wheel speed sensor.}

%The current wheel speed is calculated by Equation \ref{equa_period_v} if a valid average pedalling period $T = \bar{T}$ is extracted from the data without anomalies. 
%In practice, we apply a 5-seconds sliding window with a 1-second step to the forward accelerations for pseudo wheel speed estimation.

\section{Evaluation}\label{sec:evaluation}

In this section, we describe detailed experiment settings and conduct evaluations based on real-world riding data collected by the DiDi ride-hailing platform. 
%In addition, we compare different models, features, and demonstrate the effectiveness of each component through ablation study. 
%We also explore the riding pattern related task performance and analyze metrics such as model parameter sizes and computational efficiency to validate its feasibility.

\subsection{Implementation \& Dataset}

Existing open-source inertial data sets provide insufficient coverage for bicycle motions.
To validate the proposed solution, we construct multiple riding data sets covering different cities, users, and bicycles.
The detailed dataset include:

% To validate the proposed solution with high-precision ground truth data, we construct a customized bike experiment platform and collect data in Beijing by ourselves. The platform integrates the TDK-InvenSense ICM-42680 inertial measurement unit \footnote{https://invensense.tdk.com/products/motion-tracking/6-axis/} to capture 3-axis acceleration and 3-axis angular rate signals. Besides, it incorporates a NovAtel PP7-E1 \footnote{https://novatel.com/products/gnss-inertial-navigation-systems/combined-systems/pwrpak7-e1} high-precision navigation system to provide ground truth, which fuses GNSS with IMU and is supplemented by differential services. It achieves centimetre-level horizontal localization accuracy and about 0.5° orientation accuracy in normal riding environments, which is sufficient to serve as a reliable ground truth reference. Figure \ref{fig_bike_novatel} presents the NovAtel device and the configuration of the customized platform. During the preparation phase, the NovAtel device is strictly oriented and fixed within the bike basket, maintaining a distance of less than 20 centimetres from the IMU to ensure both reference and inertial measurements correspond to identical motion sequences.

%In addition to the customized platform, we further collect large-scale crowdsourced riding data in Hangzhou and Beijing by the DiDi ride-hailing platform to evaluate the robustness and generalization of the solution in real-world riding environments.

\begin{figure}[t]
\label{fig_bike_novatel}
    \begin{minipage}[b]{0.49\linewidth}
    % \vspace{1pt}
    \centerline{\includegraphics[width=\textwidth]{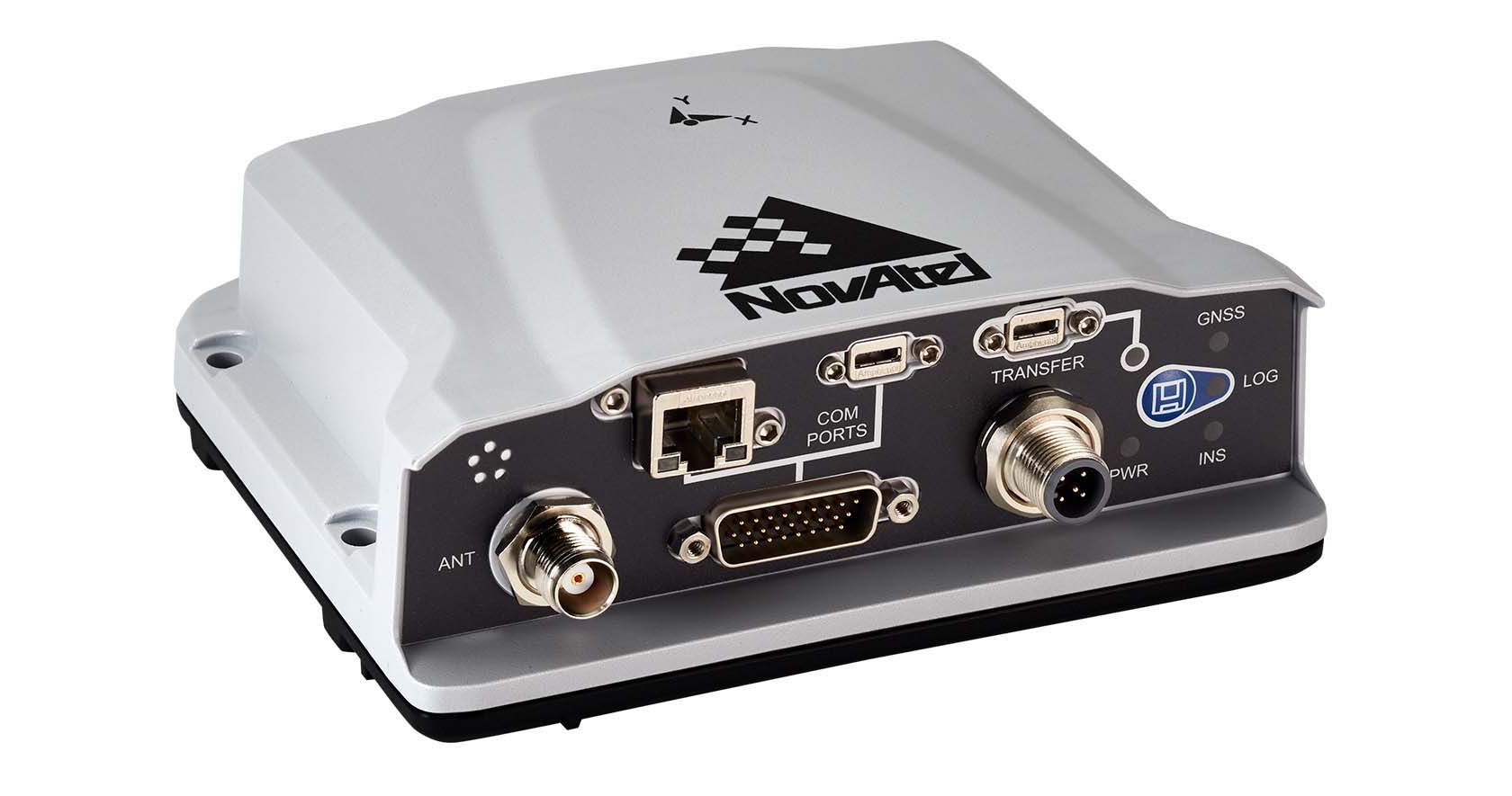}}
    \centerline{(a) NovAtel PP7-E1}
    \end{minipage}
    \begin{minipage}[b]{0.49\linewidth}
    % \vspace{1pt}
    \centerline{\includegraphics[width=\textwidth]{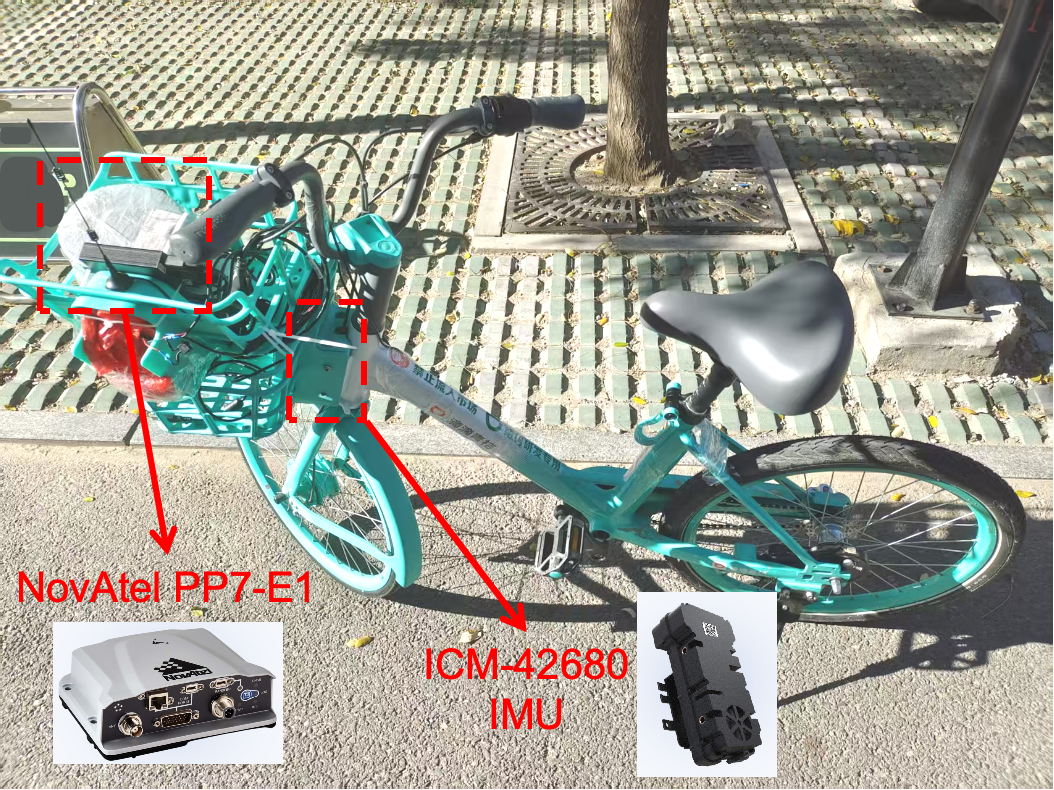}}
    \centerline{(b) Platform setup}
    \end{minipage}
\caption{NovAtel PP7-E1 device and its configuration on a bike platform for ground truth collection.}
\end{figure}

\begin{figure}[t]
\label{fig_areas}
    \centering
    \begin{minipage}{0.9\linewidth}
    % \vspace{3pt}
    \centerline{\includegraphics[width=\textwidth]{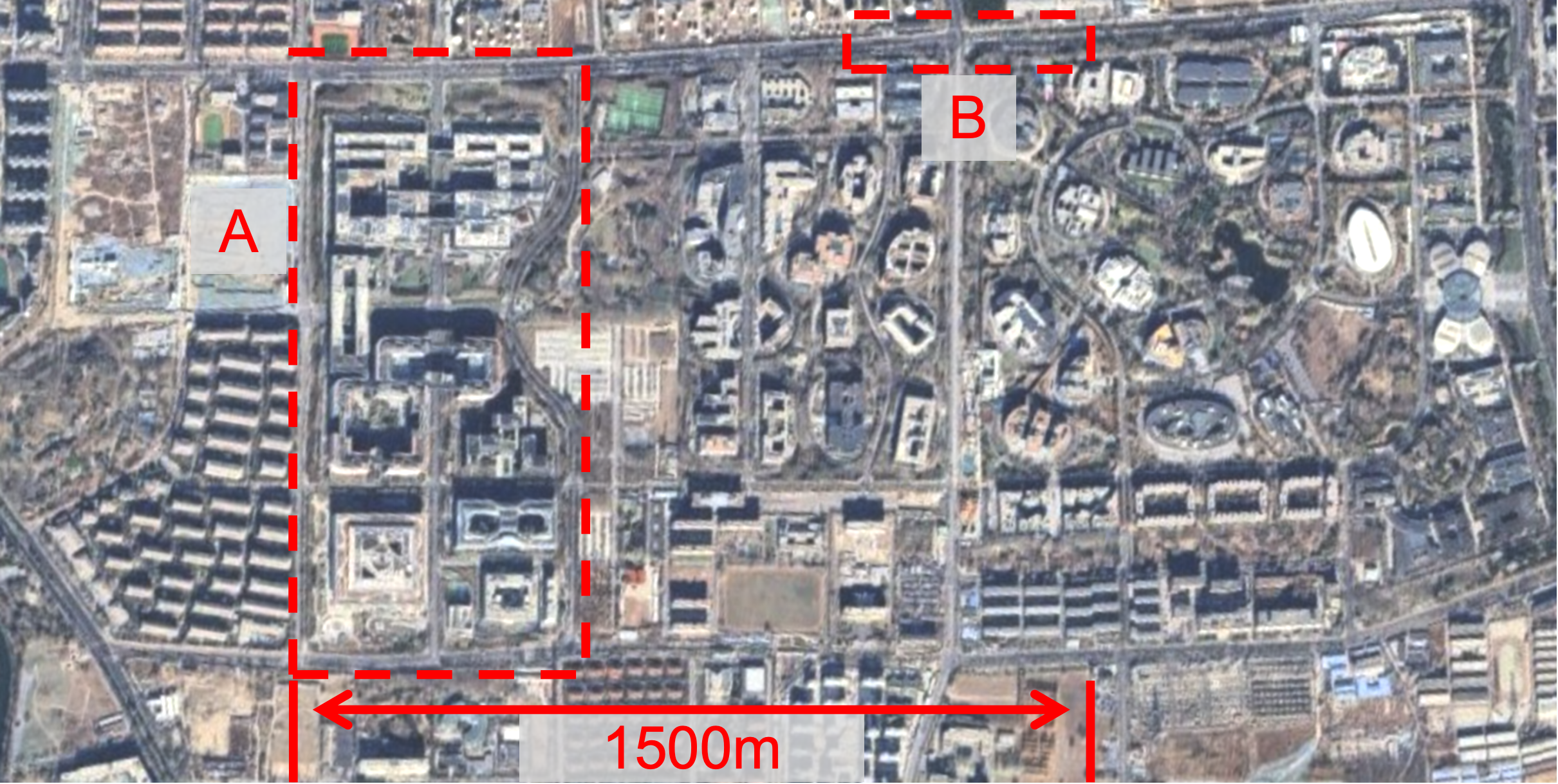}}
    \end{minipage}
    \caption{Top view of the Customized Dataset collection area.}
\end{figure}

%\subsection{Dataset}
% Existing open-source inertial data sets lack bike-specific data. Therefore, we construct riding data sets in real-world environments to validate the effectiveness of the framework. During training and evaluation of the model, we traverse each sequence with a sliding window with a length and step size of 100 to divide the inertial data. 
% The details are as follows:

\subsubsection{\textbf{Customized Dataset}} 
We construct a customized bicycle platform and collect a Customized Dataset with high-precision ground truth in Beijing. 
\rev{The platform integrates a TDK-InvenSense ICM-42680 IMU\footnote{https://invensense.tdk.com/products/motion-tracking/6-axis/} and a NovAtel PP7-E1 device\footnote{https://novatel.com/products/gnss-inertial-navigation-systems/combined-systems/pwrpak7-e1}, as shown in Figure~\ref{fig_bike_novatel}. 
The IMU records 3-axis acceleration and angular rate, while the NovAtel device provides high-precision ground truth enhanced by differential positioning services.}
The platform achieves centimetre-level horizontal localization accuracy and about 0.5 degrees orientation accuracy, making it a reliable ground truth reference. 
To ensure consistent motion measurements, the NovAtel device is rigidly mounted in the bike basket within 20 $cm$ of the IMU.

% raw
% Figure~\ref{fig_areas} marks the data collection areas, covering different trajectories, environments, and times.
% Specifically, we collect the training, validation, and test-seen sets in Area A, and build the test-seen and test-unseen sets in Area B.
% The bike moved at varying speeds during data collection, with up to about 5 $m/s$. 
% This dataset contains about 400 trajectories, each lasting 2 to 3 minutes, with a maximum distance of up to about 550 metres and a total distance of about 102 $km$.
% We divided the whole data into training, validation, test-seen and test-unseen sets with a ratio of 9:2:2:2.

Figure~\ref{fig_areas} marks the data collection areas.
\revtwo{We construct the training, validation and test-seen sets in Area A, and the test-unseen set in Area B, to evaluate model performance and its ability to generalize to unseen environments.}
The bike moved at varying speeds during data collection, with up to about $5 m/s$. 
This dataset contains about 400 trajectories, each lasting 2 to 3 minutes, with a maximum distance of up to about 550 metres and a total distance of about $102 km$. 
\revtwo{The split unit for the Customized Dataset is the complete trajectory. 
We divide all data into training, validation, test-seen and test-unseen sets based on riding distance, in a ratio of approximately 9:2:2:2.}

\subsubsection{\textbf{Crowdsourced Dataset}}  
Besides the customized platform, we further collect crowdsourced riding data in Hangzhou and Beijing by the DiDi ride-hailing platform for about two weeks, to evaluate the robustness and generalization of the solution in real-world riding environments. 
Specifically, we have collected inertial measurements and GNSS signals during riding on six bikes. 
The dataset includes trajectories from about 100 different users, with a total duration and distance of about 30 hours and 210 $km$, respectively.
\revtwo{For the Crowdsourced Dataset, the split unit is also the complete trajectory.
The test-seen set contains trajectories from areas close to those in the training data, while the test-unseen set contains trajectories from geographically distinct areas.
Similarly, we divide all data according to riding distance into training, validation, test-seen and test-unseen sets, in a ratio of approximately 39:8:13:11.}

\begin{table}[t]
  \scriptsize
  \centering
  \label{tab_period_dataset}%
  \caption{Information statistics for volunteers of the Periodic Dataset.}
  \resizebox{1.0\linewidth}{!}{ % 将表格缩放到栏宽  
    \begin{tabular}{c|c|c|c}
    \hline
    Feature & Male (n=18) & Female (n=12) & Total (n=30) \\
    \hline
    Age   & 29.89$\pm$14.49 & 36.83$\pm$13.70 & 32.67$\pm$13.21 \\
    \hline
    Height & 174.06$\pm$7.07 & 163.42$\pm$4.48 & 169.80$\pm$8.06 \\
    \hline
    Weight & 71.42$\pm$8.81 & 58.33$\pm$5.51 & 66.18$\pm$9.98 \\
    \hline
    BMI   & 23.63$\pm$3.13 & 21.88$\pm$2.30 & 22.93$\pm$2.91 \\
    \hline
    \end{tabular}%
  }
\end{table}%

\begin{table}[t]
  \Large
  \centering
  \caption{Data distribution statistics for the Periodic Dataset}
  \label{tab_period_distribution}
  \resizebox{1.0\linewidth}{!}{
  \begin{tabular}{c|c|c|c|c}
    \hline
    Platform
    & Scenarios 
    & \makecell{Mean $\pm$ Std\\ (m/s)}
    & \makecell{Total duration\\(min)}
    & \makecell{Total distance\\(km)} \\
    \hline
    \multirow{5}{*}{DiDi} 
           & Slow        & 2.282 $\pm$ 0.552 & 54.63 & 7.27 \\
           & Moderate    & 3.062 $\pm$ 0.654 & 42.50 & 6.99 \\
           & Fast        & 4.149 $\pm$ 1.183 & 18.33 & 4.29 \\
           \cline{2-5}
           & Free        & 3.357 $\pm$ 0.946 & 94.27 & 19.31 \\
           \cline{2-5}
           & \makecell{w/ Coasting} & 2.824 $\pm$ 0.649 & 10.63 & 1.75 \\
           & \makecell{w/ Stopping} & 1.869 $\pm$ 1.120 & 14.85 & 1.71 \\
           & w/ Resistance  & 3.104 $\pm$ 0.392 & 20.62 & 3.64 \\
    \hline
    Meituan & Free       & 3.299 $\pm$ 0.349 & 79.93 & 14.95 \\
    \hline
    Hello   & Free       & 3.451 $\pm$ 0.633 & 84.00 & 15.46 \\
    \hline
    Total &  &  & 419.09 & 75.37 \\
    \hline

  \end{tabular}
  }
\end{table}

% 
% \begin{table}[t]
%   \Large
%   \centering
%   \label{tab_period_distribution}%
%   \caption{Data distribution statistics for the Periodic Dataset}
%   \resizebox{1.0\linewidth}{!}{ % 将表格缩放到栏宽  
%     \begin{tabular}{c|c|c|c|c|c|c|c}
%     \hline
%     Scenarios & Slow & Moderate & Fast & Free & \makecell{w/\\ Coasting} & \makecell{w/\\ Stopping} & total \\
%     \hline
%     Mean (m/s) & 2.282 & 3.062 & 4.149 & 3.357 & 2.824 & 1.869 &  \\
%     \hline
%     Std  & 0.552 & 0.654 & 1.183 & 0.946 & 0.649 & 1.120 &  \\
%     \hline
%     Max (m/s) & 4.049 & 4.752 & 7.809 & 6.898 & 4.324 & 3.907 &  \\
%     \hline
%     \makecell{Total data \\ duration (min)} & 54.63 & 42.50  & 18.33 & 94.27 & 10.63 & 14.85 & 235.21 \\
%     \hline
%     \makecell{Total distance\\ (km)} & 7.27 & 6.99 & 4.29 & 19.31 & 1.75 & 1.71 & 41.32 \\
%     \hline
%     \end{tabular}%
%   }
% \end{table}%
%%%

\subsubsection{\textbf{Periodic Dataset}} 
To systematically evaluate the generalization and robustness of the pseudo wheel speed estimation module, we invite 30 volunteers to collect the Periodic Dataset. 
Table~\ref{tab_period_dataset} presents their age, height, weight, and Body Mass Index (BMI) as mean $\pm$ standard deviation.
Data are collected on a university campus across diverse riding scenarios, including multiple speed levels, free riding, 
\rev{riding with coasting, riding with intermittent stops, and riding with resistance under light braking.
In addition to DiDi shared bikes, we also collect free riding data from other two ride-hailing plarforms, i.e., Meituan Bike and Hello Bike, to evaluate the cross-platform reliability of the proposed PWS estimation method.}
The detailed distribution is given in Table~\ref{tab_period_distribution}.

% Affected by the resource consumption of shared bikes, it is almost impossible to use the model throughout the riding process in real scenarios, which will result in a large resource waste, and thus influence other services such as bike dispatching, maintenance, and so on. We conduct statistics on lots of different orders offline and found that the average riding time is about 10 minutes. We aim to improve the finding rate or returning rate of bikes in complex urban scenarios possible by the model. After balancing the relationship between consumption and riding duration, we choose to evaluate the performance of the model on the two-minute interval. During training, we divide the Customized Dataset according to a size of $100 \times 120$ and a step length of 100, and adopt a window of 100 to slide on each sequence. During testing, the same 100 window with step size is used.

\subsection{Baselines}

\rev{For the MTIMNet model, we compare it with one classical inertial dead reckoning baseline and several representative learning-based inertial localization methods, including IONet~\cite{Chen_Lu_Markham_Trigoni_2018}, RoNIN~\cite{9196860}, LLIO~\cite{wang2022llio}, UniTS~\cite{10.1145/3485730.3485942}, Liu \textit{et al.}~\cite{10373097}, IMUNet~\cite{10480886}, EqNIO~\cite{Jayanth25iclr}, and AirIO~\cite{11045120}. 
These baselines cover recurrent, residual, temporal-convolutional, time--frequency, and recent geometry-aware or motion-adaptive architectures for inertial motion estimation. 
For fair comparison, all learning-based baselines are trained and evaluated under the same data split and setting.}

\rev{For pseudo wheel speed (PWS) estimation, we compare it with two representative time-domain and frequency-domain periodic detection baselines respectively. 
The first is Peak Detection (PD), which estimates the pedalling period from the average interval between local peaks detected on the moving-window integration (MWI) signal. 
The second is Fast Fourier Transform (FFT)-based estimation, which transforms the MWI signal into the frequency domain and uses the dominant spectral component to infer the pedalling period.}

\begin{table*}[t]
  \centering
  \caption{Performance comparison of different models on the Customized Dataset. Red markers denote the best performance among all methods, blue markers denote the best performance among all baselines, and ``Improvement'' indicates the enhancement of red markers over blue markers.}
  \label{tab_models_evaluate}%
  \resizebox{\textwidth}{!}{%
    \begin{tabular}{c|c|c|c|c|c|c|c|c|c|c|c|cc}
    \hline
    Dataset & \makecell{Test\\ subjects} & Metric & DR    & IONet & RoNIN & LLIO  & UniTS & Liu \textit{et al.} & IMUNet & EqNIO & AirIO & MTIMNet & Improvement \\
    \hline
    \multirow{8}{*}{\makecell{Customized\\Dataset}} & \multirow{4}{*}{Seen} & ATE   & 72.16 & 20.01 & 18.18 & 17.56 & 17.83 & 38.31 & 19.79 & 17.36 & \textcolor{blue}{\textbf{17.20}} & \textcolor{red}{\textbf{14.91}} & \textbf{13.31$\% \uparrow$} \\
    \cline{3-14}          &       & RTE   & 83.73 & 20.42 & 20.03 & 19.02 & \textcolor{blue}{\textbf{17.89}} & 44.28 & 19.53 & 17.93 & 17.94 & \textcolor{red}{\textbf{14.23}} & \textbf{20.46$\% \uparrow$} \\
    \cline{3-14}          &       & PDE   & 0.41  & 0.10  & 0.09  & \textcolor{blue}{\textbf{0.08}} & \textcolor{blue}{\textbf{0.08}} & 0.20 & 0.10 & \textcolor{blue}{\textbf{0.08}} & \textcolor{blue}{\textbf{0.08}} & \textcolor{red}{\textbf{0.07}} & \textbf{12.50$\% \uparrow$} \\
    \cline{3-14}          &       & AYE   & 11.52 & 4.84  & 3.75  & 3.04 & 3.43  & 11.57 & 3.44 & \textcolor{blue}{\textbf{3.02}} & 3.41 & \textcolor{red}{\textbf{2.03}} & \textbf{32.78$\% \uparrow$} \\
    \cline{2-14}          & \multirow{4}{*}{Unseen} & ATE   & 63.02 & 18.45 & 17.18 & 15.85 & 13.82 & 39.54 & 16.15 & 13.43 & \textcolor{blue}{\textbf{13.26}} & \textcolor{red}{\textbf{11.46}} & \textbf{13.57$\% \uparrow$} \\
    \cline{3-14}          &       & RTE   & 64.63 & 21.33 & 20.85 & 17.96 & \textcolor{blue}{\textbf{14.28}} & 44.33 & 18.18 & 14.76 & 15.16 & \textcolor{red}{\textbf{12.27}} & \textbf{14.08$\% \uparrow$} \\
    \cline{3-14}          &       & PDE   & 0.28  & 0.10  & 0.08  & 0.07  & \textcolor{blue}{\textbf{0.06}} & 0.19 & 0.07 & \textcolor{blue}{\textbf{0.06}} & \textcolor{blue}{\textbf{0.06}} & \textcolor{red}{\textbf{0.05}} & \textbf{16.67$\% \uparrow$} \\
    \cline{3-14}          &       & AYE   & 11.23 & 5.90  & 5.30  & 3.43  & 2.41 & 11.28 & 3.14 & \textcolor{blue}{\textbf{2.32}} & 3.34 & \textcolor{red}{\textbf{1.78}} & \textbf{23.28$\% \uparrow$} \\
    \hline
    \end{tabular}%
  }
\end{table*}%

\begin{figure*}[t]
\label{fig_ate_aye}
    \centering
    \begin{minipage}{0.9\linewidth}
    % \vspace{3pt}
    \centerline{\includegraphics[width=\textwidth]{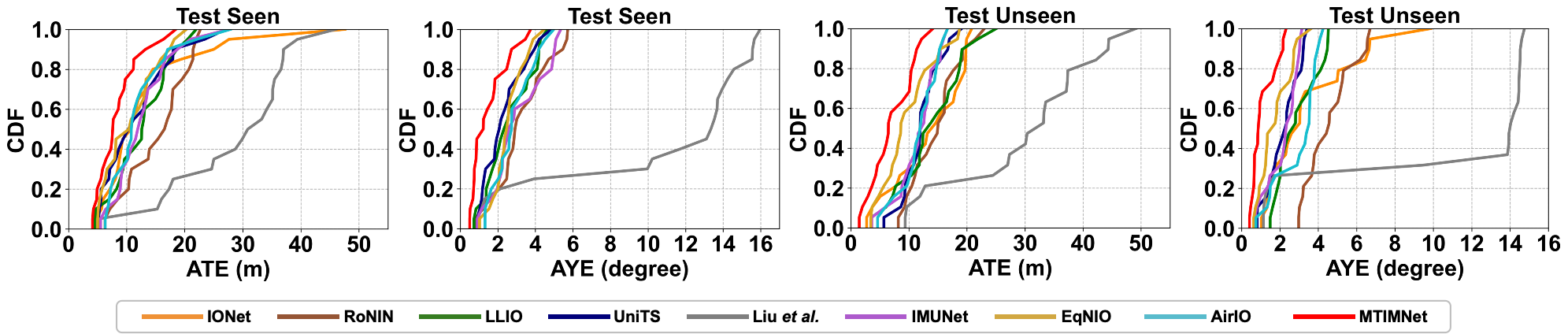}}
    \end{minipage}
    \caption{ATE and AYE for different methods on the Customized Dataset. The MTIMNet model outperforms baselines on both test-seen and test-unseen sets.}
\end{figure*}

\begin{figure*}[t]
\label{fig_trajectories}
    \centering
    \begin{minipage}{0.9\linewidth}
    % \vspace{3pt}
    \centerline{\includegraphics[width=\textwidth]{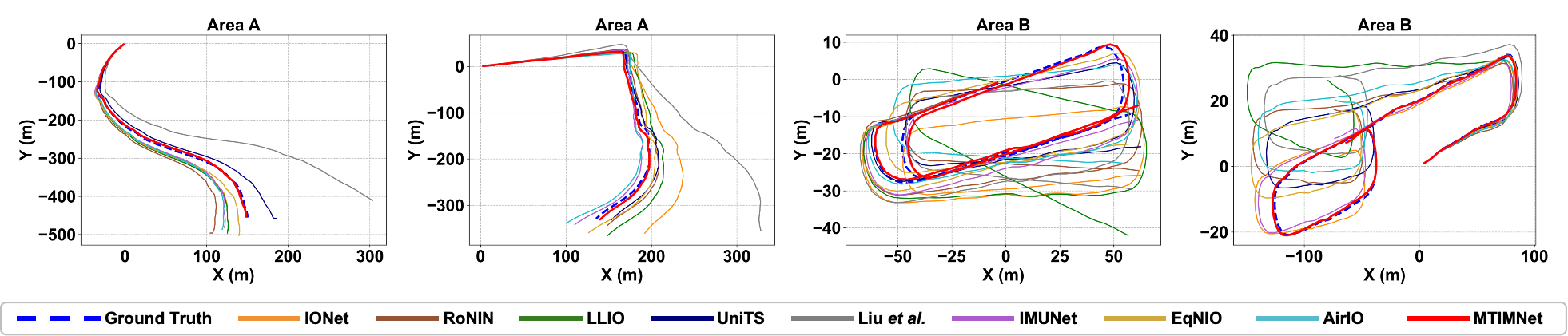}}
    \end{minipage}
    \caption{Example trajectories produced by different models. Results from the MTIMNet model are closer to the ground truth.}
\end{figure*}

\subsection{Evaluation Metrics}

\rev{We evaluate both localization accuracy and pseudo wheel speed estimation quality respectively. 
For localization, we adopt four standard metrics: Absolute Trajectory Error (ATE)~\cite{6385773}, Relative Trajectory Error (RTE)~\cite{6385773}, Position Drift Error (PDE)~\cite{rao2022ctin}, and Absolute Yaw Error (AYE)~\cite{wang2022llio}, which measure global trajectory accuracy, local trajectory consistency, drift level, and heading accuracy, respectively. 
For PWS estimation, we use Circular Error Probable (CEP) to measure the deviation between the estimated wheel speed and the high-precision speed provided by the NovAtel device.}

\subsection{Performance Comparison}

\noindent \textbf{Evaluation on the Customized Dataset:}
Table~\ref{tab_models_evaluate} reports the quantitative comparison of all models on the Customized Dataset. 
MTIMNet achieves the best performance on both the test-seen and test-unseen sets. 
\rev{Compared with the best baseline, it improves ATE and RTE by over 13\% and 14\%, respectively.}
Furthermore, a PDE of 0.07 on the test-seen set means an average error of about 7 $cm$ per meter traveled. 
In addition, the MTIMNet improves heading estimation accuracy through adaptive expert fusion and multi-task supervision, \rev{achieving at least a 23\% improvement in AYE.}

% old
% Table \ref{tab_models_evaluate} presents the quantitative comparison of all models on the Customized Dataset.
% We observe that the MTIMNet model achieves better performance on both the test-seen and test-unseen sets.
% Compared to the best baselines (e.g., UniTS and LLIO), MTIMNet reduces the errors in ATE and RTE by approximately 15\% and 14\%, respectively.
% Taking the PDE of the test-seen set as an example, 0.07 denotes that the model produces an average error of approximately 7 centimetres for every metre the bicycle travels, which is considerable in riding.
% Furthermore, MTIMNet further improves the accuracy of heading estimation through the adaptive expert fusion mechanism and the multi-task loss supervision strategy, leading to an about 30\% reduction in AYE.

Figure~\ref{fig_ate_aye} illustrates the Cumulative Distribution Function (CDF) of ATE and AYE for different models.
We observe that the MTIMNet model outperforms baseline methods on both the test-seen and test-unseen sets, thus demonstrating its robustness and accuracy.
Figure~\ref{fig_trajectories} shows visualisations of several estimated trajectories derived from real-world riding data. 
The results show that the trajectories estimated by MTIMNet are closer to the ground truth, whereas baseline methods generally fail to maintain a good balance between trajectory and heading accuracy.

\begin{table*}[t]
  \centering
  \caption{Quantitative comparison of different models on the Crowdsourced Dataset. Red values denote the best performance among all methods, while blue values denote the best performance among all baselines.}
  \label{tab_eval_on_crowdataset}%
  \resizebox{\textwidth}{!}{%
    \begin{tabular}{c|c|c|c|c|c|c|c|c|c|c|c|cc}
    \hline
    Dataset & \makecell{Test\\ subjects} & Metric & DR    & IONet & RoNIN & LLIO  & UniTS & Liu \textit{et al.} & IMUNet & EqNIO & AirIO & MTIMNet & Improvement \\
    \hline
    \multicolumn{1}{c|}{\multirow{8}{*}{\makecell{Crowdsourced\\ Dataset}}} & \multirow{4}{*}{Seen} & ATE   & 50.60  & 24.72 & 21.75 & 23.66 & 21.77 & 29.01 & 21.83 & \textcolor{blue}{\textbf{19.21}} & 19.76 & \textcolor{red}{\textbf{16.83}} & \textbf{12.39$\% \uparrow$} \\
    \cline{3-14}          &       & RTE   & 59.34 & 27.72 & 24.04 & 26.89 & 25.07 & 31.20 & 24.23 & \textcolor{blue}{\textbf{21.74}} & 22.66 & \textcolor{red}{\textbf{18.57}} & \textbf{14.58$\% \uparrow$} \\
    \cline{3-14}          &       & PDE   & 0.61  & 0.22  & 0.19  & 0.25  & \textcolor{blue}{\textbf{0.16}} & 0.38 & 0.24 & 0.21 & 0.34 & \textcolor{red}{\textbf{0.14}} & \textbf{12.50$\% \uparrow$} \\
    \cline{3-14}          &       & AYE   & 17.74 & 12.35 & \textcolor{blue}{\textbf{9.12}} & 14.30  & 10.77 & 17.81 & 9.38 & 11.29 & 9.45 & \textcolor{red}{\textbf{7.75}} & \textbf{15.02$\% \uparrow$} \\
    \cline{2-14}          & \multirow{4}{*}{Unseen} & ATE   & 44.91 & 22.95 & 21.58 & 26.82 & \textcolor{blue}{\textbf{19.74}} & 27.72 & 21.42 & 20.22 & 19.78 & \textcolor{red}{\textbf{16.26}} & \textbf{17.63$\% \uparrow$} \\
    \cline{3-14}          &       & RTE   & 51.60  & 25.39 & 23.81 & 29.19 & 22.71 & 30.17 & 23.40 & 22.08 & \textcolor{blue}{\textbf{21.13}} & \textcolor{red}{\textbf{18.13}} & \textbf{14.20$\% \uparrow$} \\
    \cline{3-14}          &       & PDE   & 0.51  & 0.26  & 0.21  & 0.28  & \textcolor{blue}{\textbf{0.19}} & 0.38 & 0.29 & 0.23 & 0.35 & \textcolor{red}{\textbf{0.15}} & \textbf{21.05$\% \uparrow$} \\
    \cline{3-14}          &       & AYE   & 27.97 & 13.74 & 12.33 & 21.41 & 11.70 & 28.09 & \textcolor{blue}{\textbf{11.10}} & 11.15 & 11.74 & \textcolor{red}{\textbf{9.54}} & \textbf{14.05$\% \uparrow$} \\
    \hline
    \end{tabular}%
  }
\end{table*}%

\noindent \textbf{Evaluation on the Crowdsourced Dataset:} 
To validate robustness in more general scenarios, we further compare MTIMNet with the baselines on the Crowdsourced Dataset, as shown in Table~\ref{tab_eval_on_crowdataset}. 
\rev{Although the baselines achieve acceptable accuracy, MTIMNet outperforms them on all metrics, with improvements of at least 12\% in ATE and 14\% in AYE.}
Moreover, while baseline errors increase on the more complex crowdsourced data, the MTIMNet maintains stable performance, demonstrating stronger robustness and generalization.

% old
% To validate the robustness of the model in general scenarios, we further compared the performance of the MTIMNet model with baselines on the Crowdsourced Dataset, as shown in Table \ref{tab_eval_on_crowdataset}.
% Although the baselines achieve acceptable accuracy, MTIMNet outperforms them on all metrics, and yields improvements of at least 17\% and 15\% in terms of ATE and AYE, respectively.
% Furthermore, the baseline methods exhibit a significant increase in errors when performed on more complex crowdsourced data.
% In contrast, MTIMNet maintains stable performance under identical conditions, which reflects its robustness and generalization.

\subsection{Model Performance Study}

\subsubsection{Task Study}
\begin{figure}[t]
    \begin{minipage}[b]{0.49\linewidth}
    % \vspace{1pt}
    \centerline{\includegraphics[width=\textwidth]{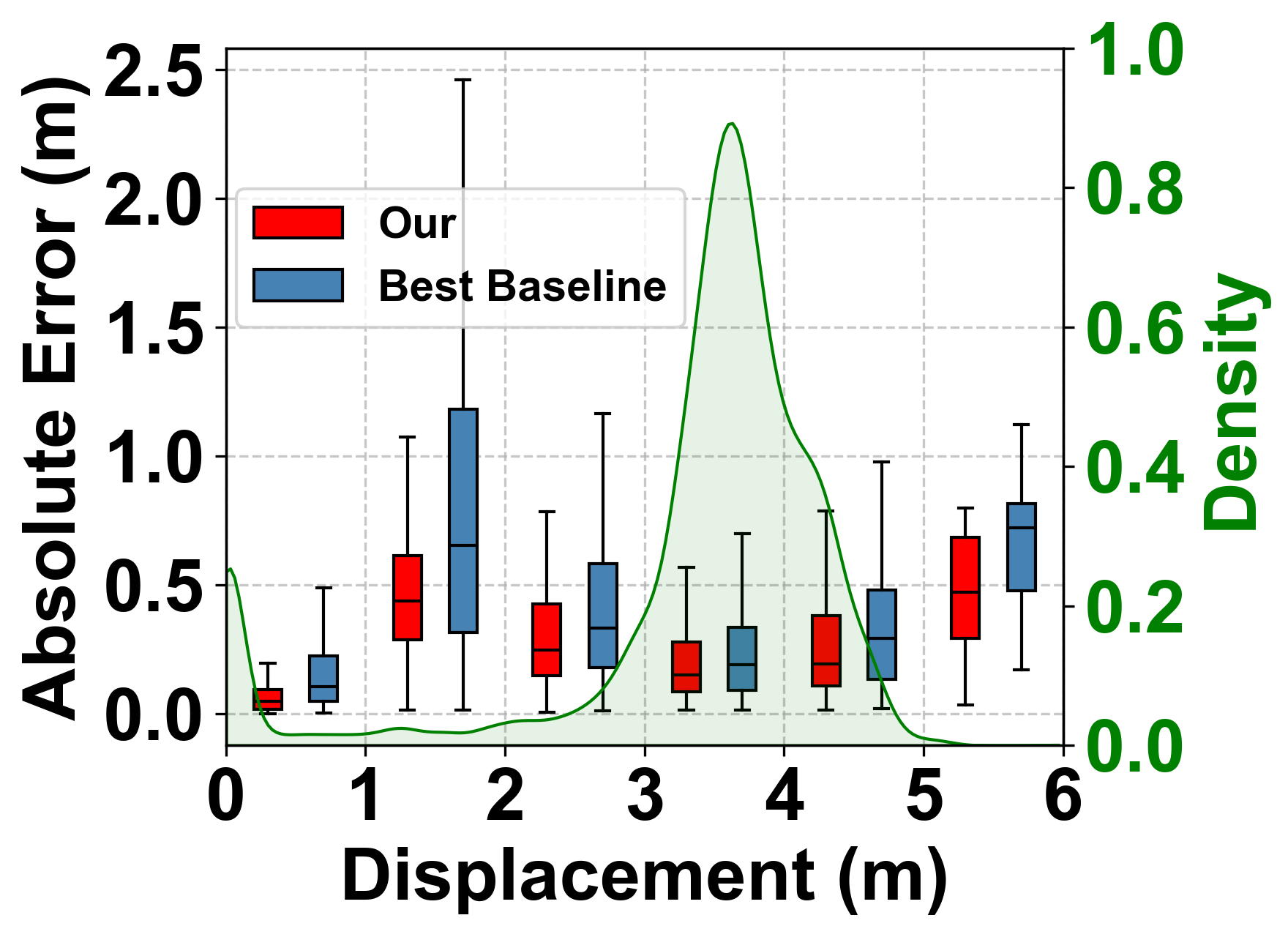}}
    \vspace{-5pt}
    \centerline{(a) Displacement errors}
    \end{minipage}
    \begin{minipage}[b]{0.49\linewidth}
    % \vspace{1pt}
    \centerline{\includegraphics[width=\textwidth]{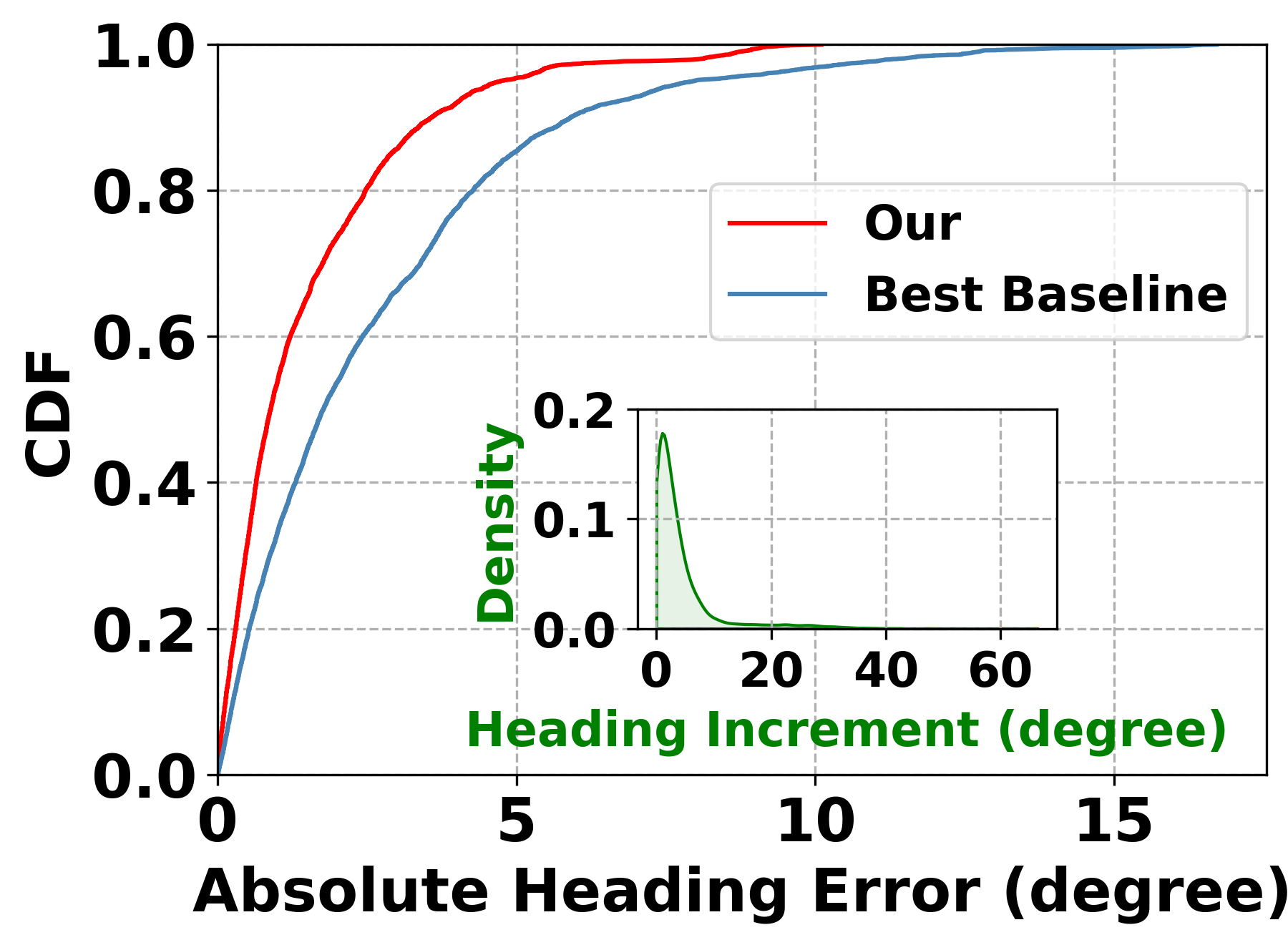}}
    \vspace{-5pt}
    \centerline{(b) Absolute heading errors}
    \end{minipage}
\caption{Comparison of MTIMNet with the best baseline in terms of displacement and heading. The green curves show the distribution of ground truth for displacement and heading increments, respectively.}
\label{fig_disp_head_err}
\end{figure}

To evaluate MTIMNet on different tasks, we compare its displacement and heading estimates with those of the best baseline, UniTS, against the ground truth. 
Figure~\ref{fig_disp_head_err}(a) shows the ground truth displacement distribution, where samples are divided into 1 metre intervals to compare displacement errors. 
\rev{MTIMNet yields lower displacement errors across all intervals, remaining almost always below 1 metre even when the riding speed exceeds $5 m/s$.}

\rev{Figure~\ref{fig_disp_head_err}(b) shows the CDF of absolute heading errors for MTIMNet and the best baseline, while the inset gives the distribution of ground truth for heading increments.}
MTIMNet achieves higher heading accuracy while maintaining strong displacement performance, indicating that its shared experts and task-specific gating effectively balance different tasks.

% old
% To evaluate the performance of the MTIMNet model on different tasks, we extract the displacement and heading increment estimates from both MTIMNet and the best baseline UniTS at each moment and compare them with ground truth. 
% Figure \ref{fig_disp_head_err}(a) illustrates the probability density distribution of the ground truth for displacement measurements.
% We divide these measurements into intervals of 1 metre and compare the errors of MTIMNet and UniTS over different intervals.
% We observe that the displacement estimation accuracy of MTIMNet outperforms UniTS across all intervals.
% Furthermore, the absolute displacement errors of MTIMNet remain almost always below 1 metre at any riding speed, even at maximum velocity exceeding 5 $m/s$.

% Similarly, Figure \ref{fig_disp_head_err}(b) presents the CDF of the heading increment estimation errors, and the probability density distribution of the ground truth for heading increments.
% As shown in the figure, MTIMNet achieves greater heading increment estimation accuracy compared to the UniTS while maintaining displacement estimation precision, with an improvement of approximately 0.93 degrees per second. 
% Thus, the shared experts and task-specific gating mechanisms of MTIMNet effectively balance the performance across different tasks, yielding robust estimates.

\subsubsection{Model Structure Study}
In MTIMNet, the encoder extracts basic features from raw IMU sequences for the experts and gating modules. 
\rev{To evaluate different architectures, we compare encoders built with fully connected layers (FC), convolutional neural networks (CNN), and long short-term memory networks (LSTM), and also test a scheme that directly uses handcrafted time- and frequency-domain features from IMU readings without an encoder.}
% old
% In MTIMNet, the encoder extracts basic motion features from raw IMU sequences, and these features serve as input to the experts and gating modules.
% To explore the effect of different structures, we build the encoder with modules such as Fully Connected (FC) layers and the Long Short-Term Memory (LSTM) network, respectively, and compare them with the current structure. 
% In addition, we also attempted to extract time-frequency features directly from IMU readings as basic features, rather than employing an encoder.

\begin{figure}[t]
    \begin{minipage}[b]{0.48\linewidth}
    % \vspace{1pt}
    \centerline{\includegraphics[width=\textwidth]{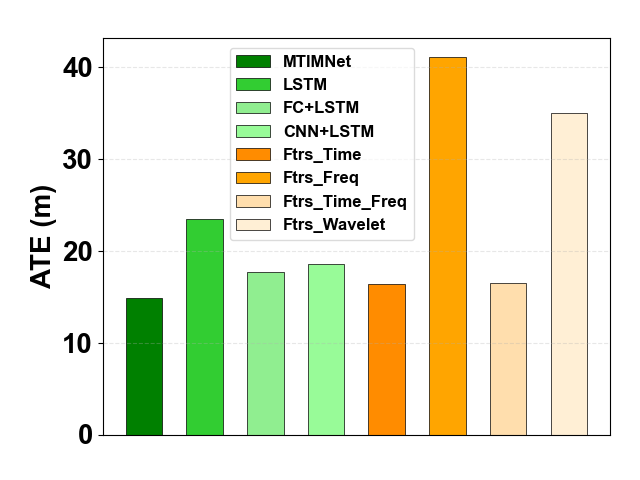}}
    \vspace{-5pt}
    \centerline{(a) ATE}
    \end{minipage}
    \hfill
    \begin{minipage}[b]{0.48\linewidth}
    % \vspace{1pt}
    \centerline{\includegraphics[width=\textwidth]{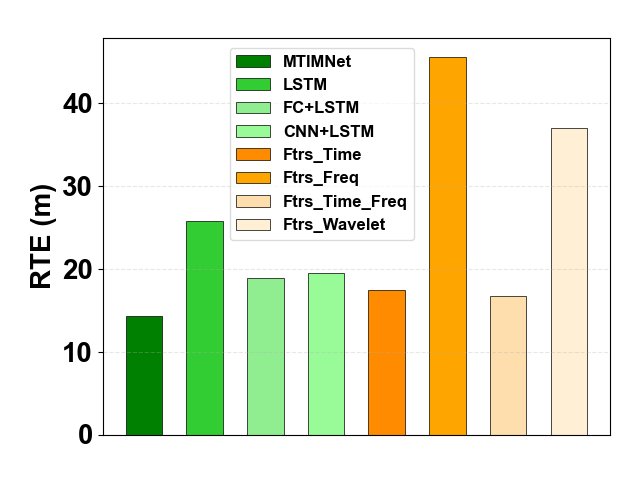}}
    \vspace{-5pt}
    \centerline{(b) RTE}
    \end{minipage}
    \begin{minipage}[b]{0.48\linewidth}
    % \vspace{1pt}
    \centerline{\includegraphics[width=\textwidth]{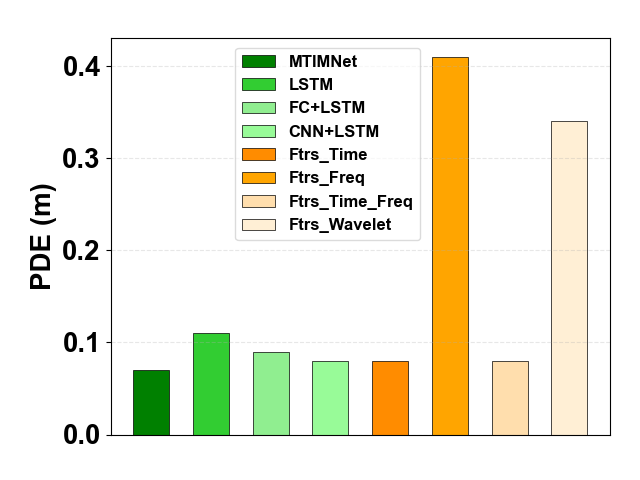}}
    \vspace{-5pt}
    \centerline{(c) PDE}
    \end{minipage}
    \hfill
    \begin{minipage}[b]{0.48\linewidth}
    % \vspace{1pt}
    \centerline{\includegraphics[width=\textwidth]{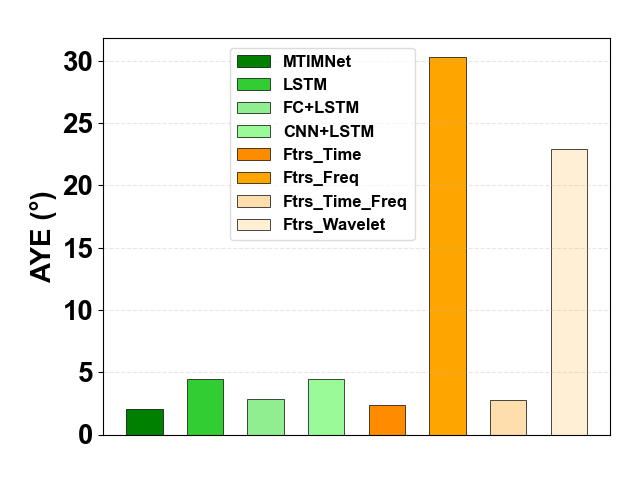}}
    \vspace{-5pt}
    \centerline{(d) AYE}
    \end{minipage}
\caption{Performance comparison of MTIMNet with different model structures or input features. Green bars: different network architectures. Orange bars: different input features, which are used to replace the encoder module.}
\label{fig_model_structure_study}
\end{figure}

Figure~\ref{fig_model_structure_study} compares these methods on different metrics. 
Here, ``LSTM'' uses an LSTM encoder, while ``FC+LSTM'' and ``CNN+LSTM'' use FC and CNN encoders with LSTM task heads. 
\rev{``Ftrs\_'' denotes handcrafted time-, frequency-, and wavelet-domain features from 6-axis IMU data\footnote{\rev{``Time'' includes maximum, minimum, mean, standard deviation, and integration. ``Freq'' includes dominant frequency, spectral entropy, energy, spectral centroid, and PSD peaks from FFT. ``Wavelet'' includes sub-band energy ratios extracted using the ``db3'' basis with three-level decomposition.}}.}
The results show that the CNN-based encoder performs best, while time-domain features are \rev{the most effective among the handcrafted alternatives.}
% old 
% Figure \ref{fig_model_structure_study} illustrates the performance of the above methods on different metrics.
% Specifically, ``LSTM'' refers to building the encoder with LSTM.
% ``FC+LSTM'' and ``CNN+LSTM'' means constructing the encoder with FC and CNN, respectively, and exploiting LSTM as task heads.
% ``Ftrs\_'' means extracting different features from the 6-axis IMU data as input. 
% Specifically, ``Time'' includes maximum, minimum, average, standard deviation, and integration.
% ``Freq'' includes the dominant frequency, spectral entropy, energy, spectral centroid, and peaks of the power spectral density extracted from the Fast Fourier Transform.
% ``Wavelet'' consists of the energy ratio features of the inertial signals in each sub-band extracted from the ``db3'' wavelet basis function and the 3-layer decomposition. 
% The results show that the CNN-based encoder achieves the best performance on all metrics. 
% Furthermore, among these features, the time-domain features play a dominant role in localization accuracy.

\subsubsection{Uncertainty Study}
\begin{figure}[t]
\label{fig_disp_head_uncer}
    \begin{minipage}[b]{0.49\linewidth}
    % \vspace{1pt}
    \centerline{\includegraphics[width=\textwidth]{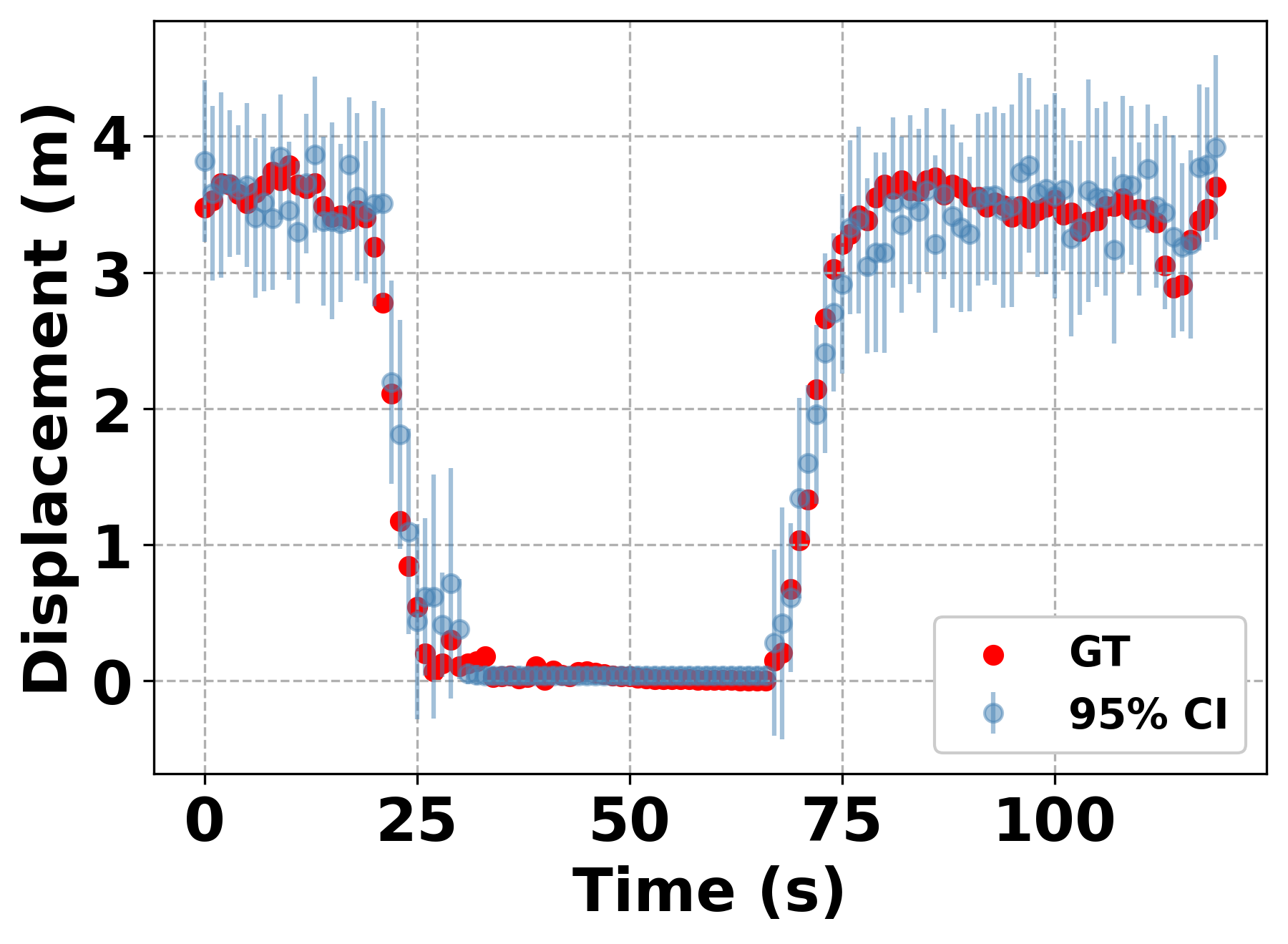}}
    \vspace{-3pt}
    \centerline{(a) Displacement}
    \end{minipage}
    \begin{minipage}[b]{0.49\linewidth}
    % \vspace{1pt}
    \centerline{\includegraphics[width=\textwidth]{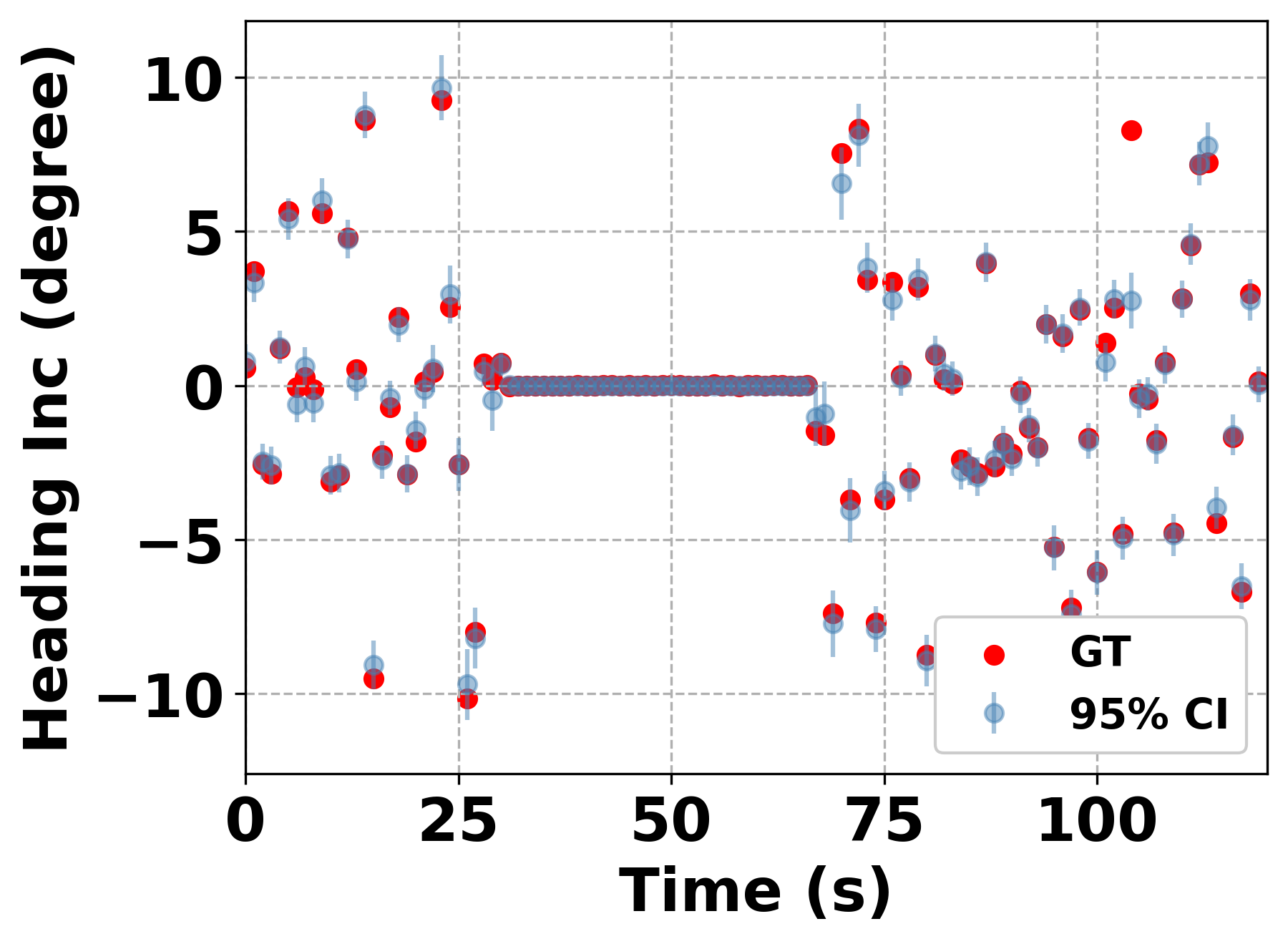}}
    \vspace{-3pt}
    \centerline{(b) Heading Increment}
    \end{minipage}
\caption{Uncertainty estimation effects for different tasks. Red dots refer to ground truth, blue dots refer to model estimates, and blue lines indicate uncertainty bounds. The ground truth is almost entirely covered.}
\end{figure}

To validate the accuracy of the uncertainty estimates, we quantitatively evaluate the coverage of the ground truth within the estimated confidence intervals. 
For displacement $\Delta d$ and heading increment $\Delta \psi$, we construct the 95\% Confidence Interval (CI) as $CI=[\hat{y}_t-z\cdot \sigma_t, \hat{y}_t+z\cdot \sigma_t]$. 
$\hat{y}_t$ denotes the estimated value of $\Delta d$ and $\Delta \psi$ at moment \rev{$t$}, $\sigma_t$ is the uncertainty estimation (standard deviation), and $z$=1.96 is the standard normal quantile at the 95\% confidence level.
Figure~\ref{fig_disp_head_uncer} demonstrates the accuracy of the estimates, where the ground truth (red dots) are almost entirely contained within the uncertainty bounds from the model (blue lines).
The MTIMNet model achieves uncertainty estimation coverage rates of 94.7\% for $\Delta d$ and 95.2\% for $\Delta \psi$, which are close to the ideal 95\% confidence level.

\begin{figure}[t]
    \begin{minipage}[b]{0.49\linewidth}
    % \vspace{1pt}
    \centerline{\includegraphics[width=\textwidth]{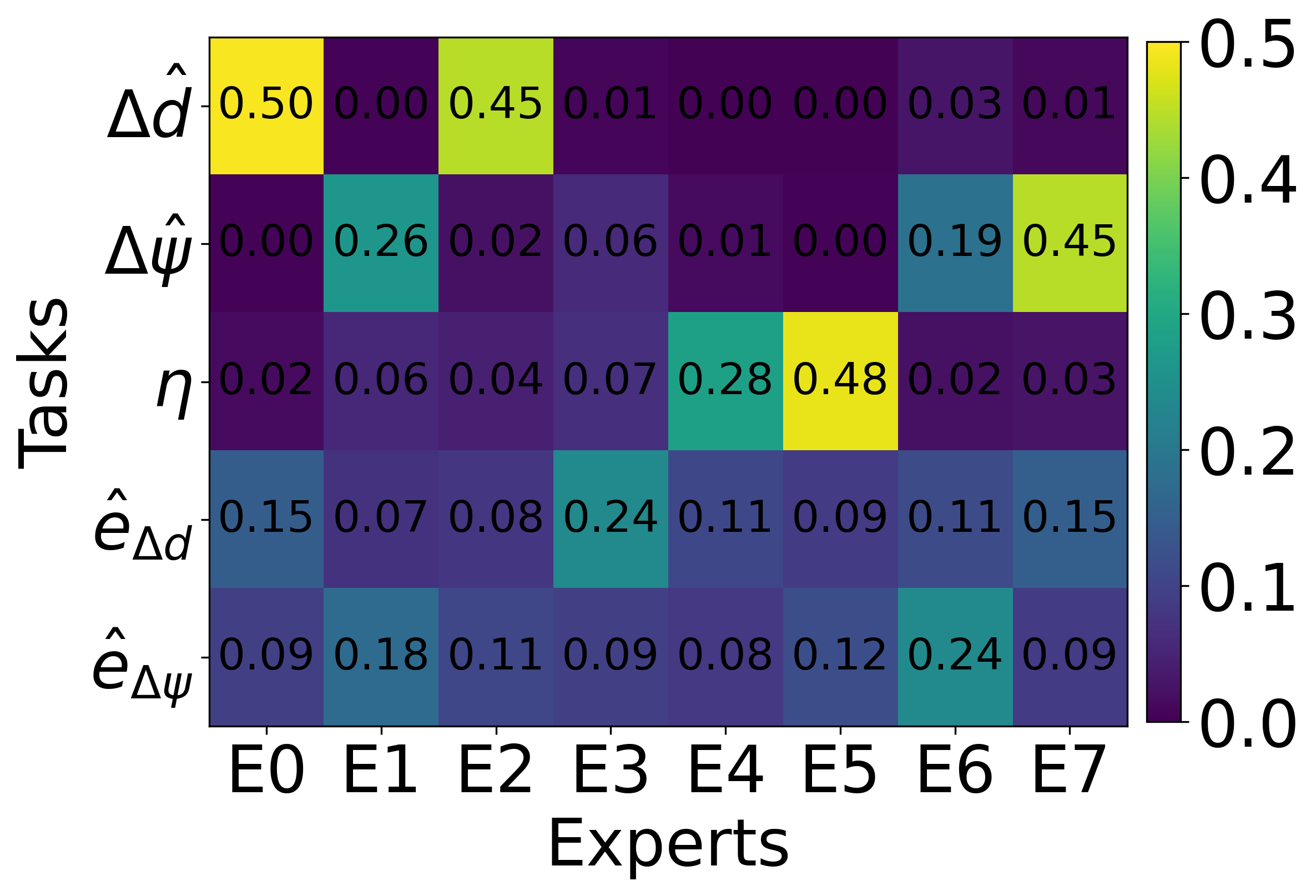}}
    \vspace{-3pt}
    \centerline{\rev{(a) Heatmap}}
    \end{minipage}
    \begin{minipage}[b]{0.49\linewidth}
    % \vspace{1pt}
    \centerline{\includegraphics[width=\textwidth]{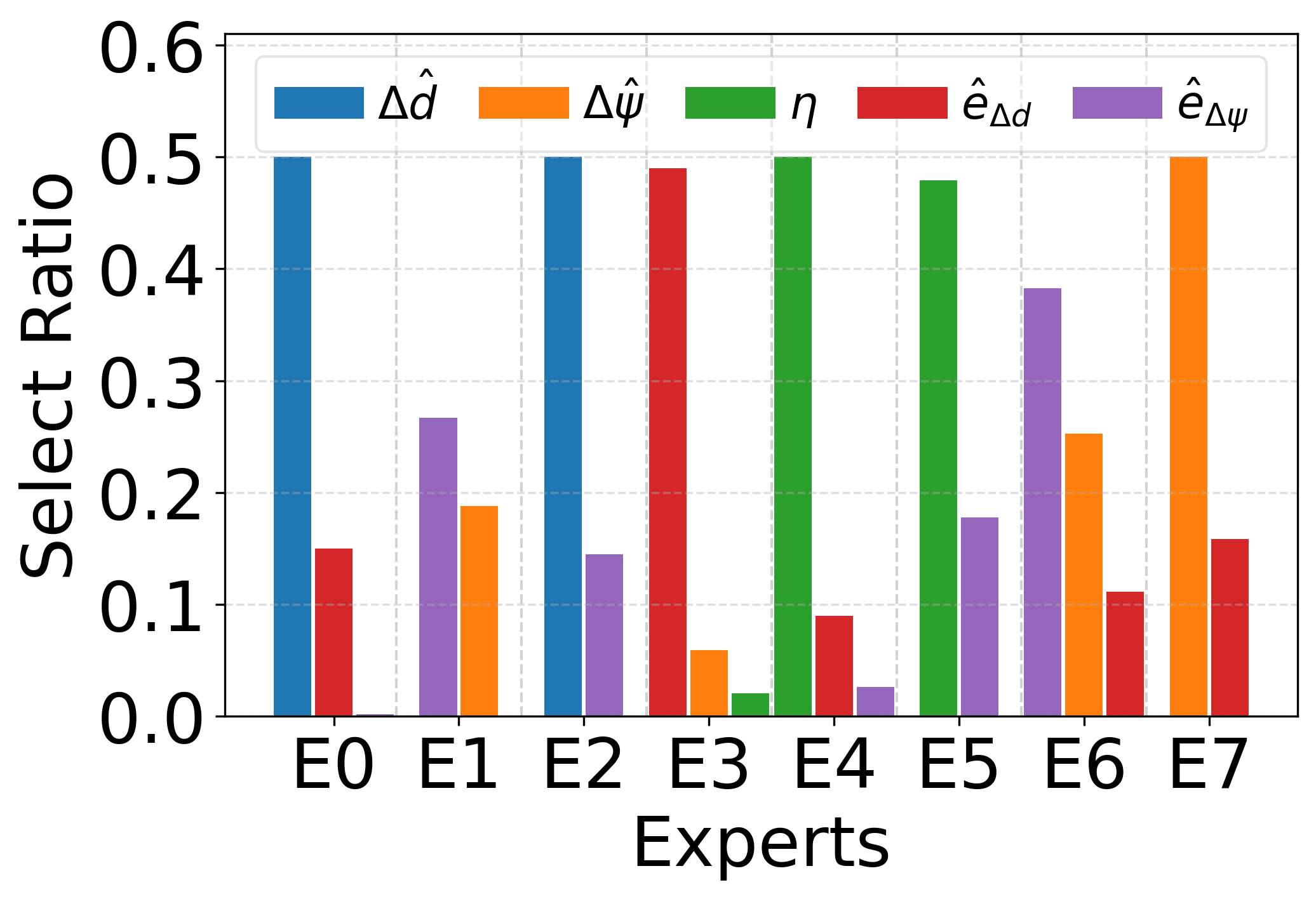}}
    \vspace{-3pt}
    \centerline{\rev{(b) Select Ratio}}
    \end{minipage}
\caption{\rev{Task-expert analysis of the proposed MTIMNet. (a) Average gating weights assigned to each expert for different tasks. (b) Expert selection ratios under sparse top-2 routing for different tasks.}
}
\label{fig_heatmap_select_ratio}
\end{figure}

\subsubsection{\rev{MoE behavior interpretability}}
\rev{To further analyze the MoE behavior, Figure~\ref{fig_heatmap_select_ratio} visualizes the expert selection patterns of different tasks. 
The heatmap shows the average gating weights, and the bar chart shows the frequency of each expert being selected in the top-2 routing results. 
Different tasks exhibit distinct expert preferences, rather than uniformly relying on the same shared representation. 
This indicates that the proposed model performs task-specific routing and captures partially shared yet task-dependent motion representations.}

\subsubsection{\rev{Experts and Top-k Study}} 
\rev{To justify the expert-routing design, we compare different numbers of experts and top-k routing strategies, as summarized in Table~\ref{tab_experts_topk}.
The best performance is achieved with 8 experts and top-2 routing. 
Fewer experts limit representation ability, whereas more experts provide no further gain. 
Compared with top-2, both top-1 and denser fusion perform worse, suggesting that sparse task-specific routing is more effective than dense shared fusion.}

\begin{table}[t]
\centering
\caption{\rev{Performance comparison under different numbers of experts and top-k routing strategies. For each setting, the first row in each cell denotes ATE, while the second row denotes AYE.}}
\label{tab_experts_topk}
\renewcommand{\arraystretch}{1.0}
\small
\begin{tabular}{c|c|c|c|c|c}
\hline
\diagbox[width=8em]{\textbf{Experts}}{\textbf{Top-k}} & \textbf{top-1} & \textbf{top-2} & \textbf{top-4} & \textbf{top-5} & \textbf{All} \\
\hline
\textbf{5}  & \makecell{16.56 \\ 2.59} 
   & \makecell{15.44 \\ 2.34} 
   & \makecell{16.01 \\ 2.43} 
   & \multicolumn{2}{c}{\makecell{17.09 \\ 3.05}} \\
\hline
\textbf{8}  & \makecell{15.31 \\ 2.57} 
   & \makecell{\textbf{14.91} \\ \textbf{2.03}} 
   & \makecell{17.61 \\ 3.01} 
   & \makecell{18.55 \\ 3.86} 
   & \makecell{19.03 \\ 3.55} \\
\hline
\textbf{10} & \makecell{15.76 \\ 2.44} 
   & \makecell{15.13 \\ 2.32} 
   & \makecell{15.97 \\ 2.52} 
   & \makecell{16.89 \\ 2.70} 
   & \makecell{17.28 \\ 3.30} \\
\hline
\end{tabular}
\end{table}

\subsection{Pseudo Wheel Speed Estimation Performance Study}
\label{sec_eval_pws}

\noindent \textbf{Evaluation on the Periodic Dataset:}
To demonstrate the effectiveness of the autocorrelation-based periodic detection method, we compared it with two baselines: the PD method and the FFT-based method, on the Periodic Dataset.
Specifically, all methods run on the MWI sequences from forward acceleration data, using a 5 seconds sliding window with a 1-second step. 
%Since our method includes an anomaly detection component based on autocorrelation, while the two comparison methods do not have such a mechanism.

Table~\ref{tab_period_method_comp} quantifies the overall performance of different periodic detection methods, where the Time Coverage Ratio (TCR) denotes the ratio of the time when pseudo wheel speed is detected to the total time. 
\revtwo{Although PD can produce estimates in most cases, there are significant errors. Our method rejects instances lacking reliable periodic features, making it more robust in practical scenarios and capable of providing reliable velocity observations for calibration.
Consequently,} 
\rev{our method achieves the best balance between estimation accuracy and valid detection coverage, showing that the periodic detection and anomaly filtering strategy is effective for extracting reliable pedal-induced speed observations.}

\begin{table}[t]
  \scriptsize
  \centering
  \label{tab_period_method_comp}%
  \caption{Comparison of different periodic detection algorithms. The TCR denotes the ratio of the time when pseudo wheel speed is detected to the total time across all data.}
  \resizebox{1.0\linewidth}{!}{ % 将表格缩放到栏宽  
    \begin{tabular}{c|c|c|c|c|c}
    \hline
    Method & \makecell{CEP68\\(m/s)} & \makecell{CEP80\\(m/s)} & \makecell{CEP95\\(m/s)} & \makecell{Max\\(m/s)}   & TCR $\uparrow$ \\
    \hline
    Ours  & \textbf{0.18} & \textbf{0.24} & \textbf{0.42} & \textbf{1.28} & 50.0\% \\
    \hline
    PD    & 0.81  & 1.09  & 1.89  & 5.36  & \textbf{98.0\%} \\
    \hline
    FFT   & 0.24  & 0.34  & 1.57  & 3.41  & 54.0\% \\
    \hline
    \end{tabular}%
  }
\end{table}%

\begin{figure}[t]
\centering
    \begin{minipage}[b]{0.49\linewidth}
    % \vspace{1pt}
    \centerline{\includegraphics[width=\textwidth]{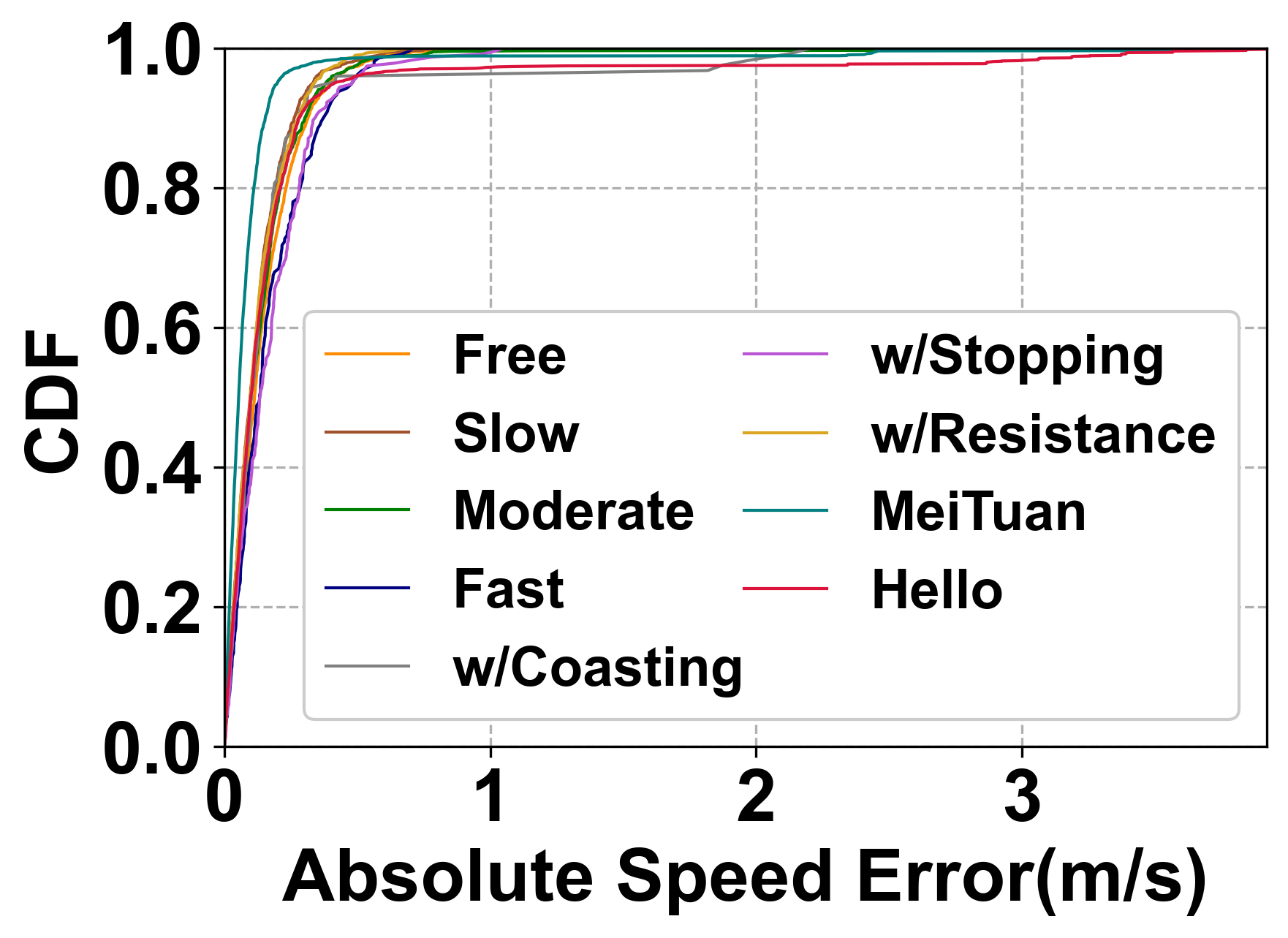}}
    \vspace{-3pt}
    \centerline{\rev{(a) Speed Error}}
    \end{minipage}
    \begin{minipage}[b]{0.49\linewidth}
    % \vspace{1pt}
    \centerline{\includegraphics[width=\textwidth]{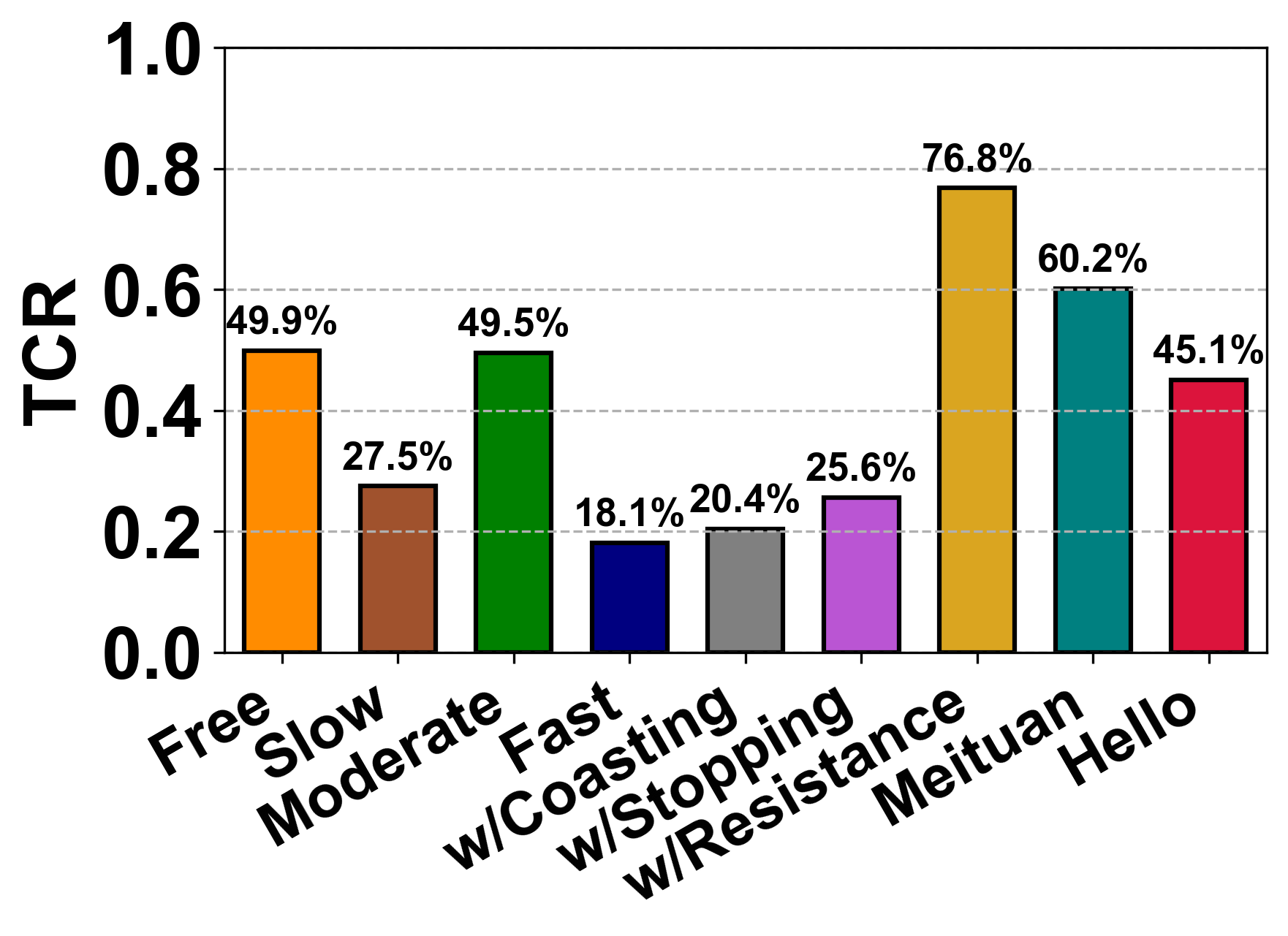}}
    \vspace{-3pt}
    \centerline{\rev{(b) TCR}}
    \end{minipage}
\caption{\rev{Performance of pseudo wheel speed estimation under different riding scenarios and bike platforms.}}
\label{fig_pws_performence}
\end{figure}

\noindent \textbf{Comparison in different scenarios \rev{and bike platforms}:} 
Figure~\ref{fig_pws_performence} summarizes the pseudo wheel speed estimation performance under different riding scenarios \rev{and across different shared bike platforms.}
The method maintains low speed error under stable pedalling, while the TCR decreases in fast, coasting, and stopping cases where pedalling periodicity is weak or absent, \rev{indicating that pseudo wheel speed is activated only under valid pedalling conditions.
The results on riding with resistance and on Meituan Bike and Hello Bike further suggest that variations in bike condition and mechanical configuration mainly affect the estimation bias and valid detection ratio, but do not invalidate the proposed periodic detection framework.}

\begin{figure}[t]
\centering
% ===== 1 =====
\begin{minipage}[b]{0.49\linewidth}
% \vspace{1pt}
\centerline{\includegraphics[width=\textwidth]{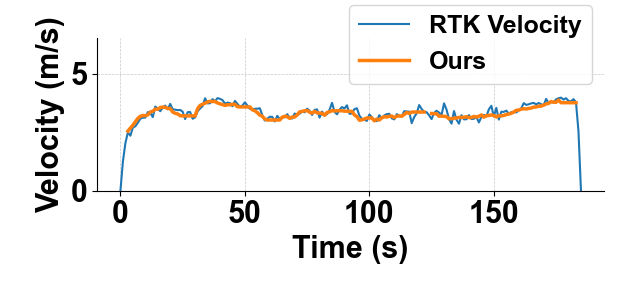}}
\vspace{-5pt}
\centerline{(a) Free}
\end{minipage}
% ===== 2 =====
\begin{minipage}[b]{0.49\linewidth}
% \vspace{1pt}
\centerline{\includegraphics[width=\textwidth]{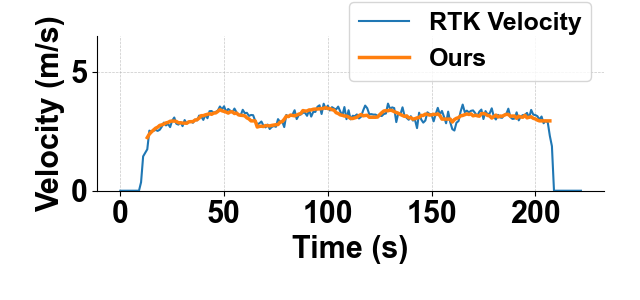}}
\vspace{-5pt}
\centerline{\rev{(b) w/ Resistance}}
\end{minipage}
% ===== 3 =====
\begin{minipage}[b]{0.49\linewidth}
% \vspace{1pt}
\centerline{\includegraphics[width=\textwidth]{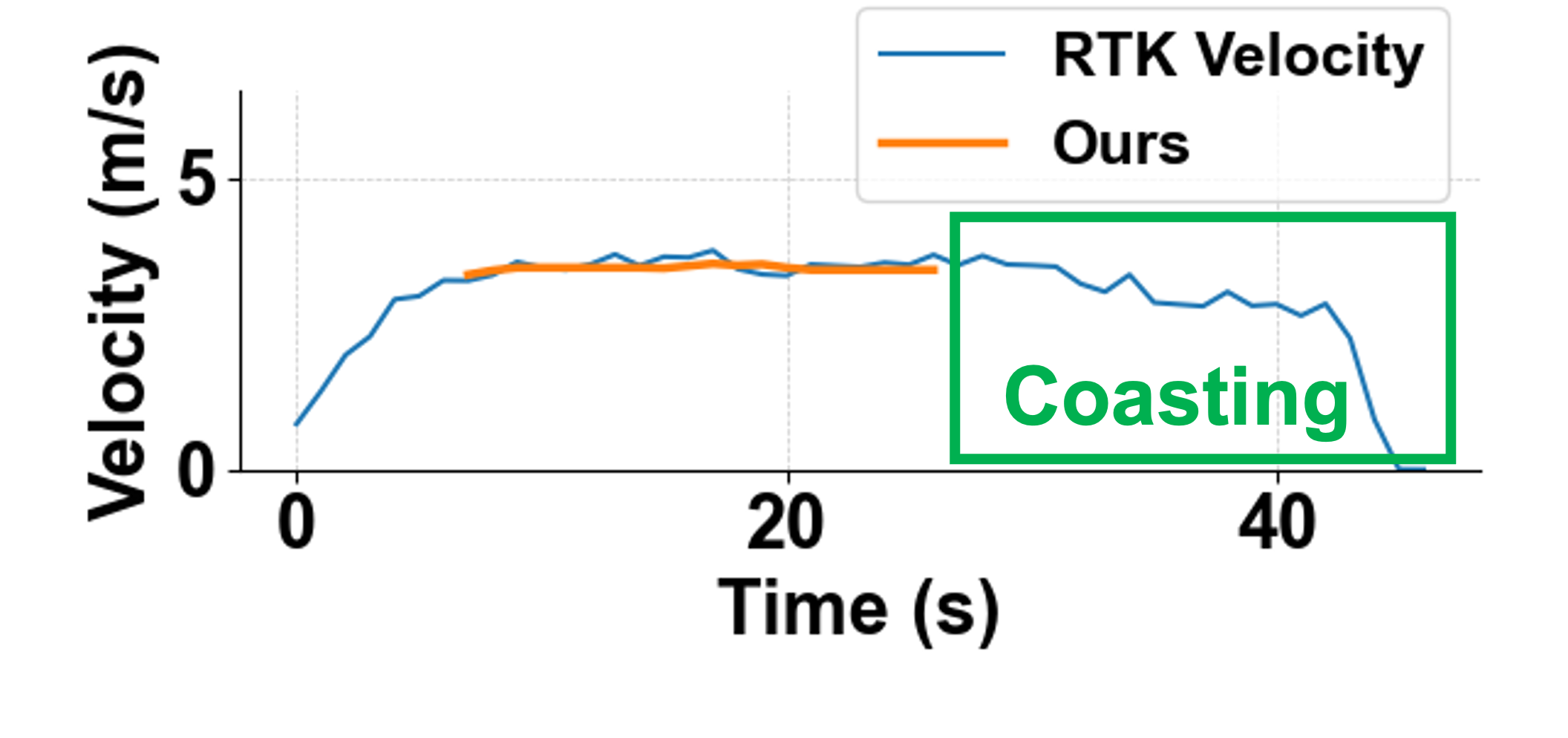}}
\vspace{-5pt}
\centerline{\rev{(c) w/ Coasting}}
\end{minipage}
% ===== 4 =====
\begin{minipage}[b]{0.49\linewidth}
% \vspace{1pt}
\centerline{\includegraphics[width=\textwidth]{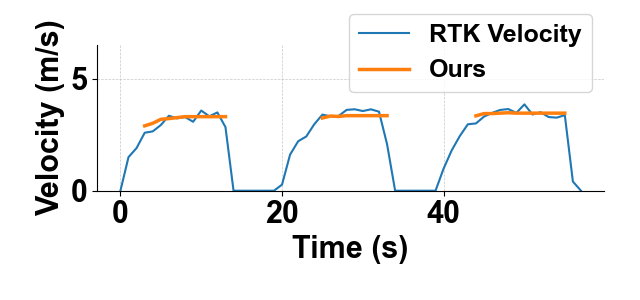}}
\vspace{-5pt}
\centerline{(d) w/ Stopping}
\end{minipage}
\caption{\rev{Comparison of pseudo wheel speed estimation with RTK velocity in four representative riding scenarios.}}
\label{fig_period_free_static}
\end{figure}

Figure~\ref{fig_period_free_static} presents the speed estimation results for \rev{four} representative scenarios, including free riding, \rev{riding with resistance, riding with coasting,} and riding with intermittent stops. 
The estimated pseudo wheel speed closely follows the RTK velocity during valid pedalling intervals, while naturally suppressing outputs during coasting or non-pedalling segments. 
\rev{These examples demonstrate that the pseudo wheel speed can provide sufficiently reliable auxiliary observations under periodic riding patterns.}

\subsection{Ablation Study}

\noindent \textbf{MTIMNet:}
We evaluate the effectiveness of different components in MTIMNet from \rev{four} aspects, including the model structure, \rev{residual estimation, uncertainty estimation}, and multi-task loss. 
\rev{Table~\ref{tab_ablation_study} presents the complete comparison.}

% \begin{figure}[t]
% \label{fig_ablation_study}
%     \centering
%     \begin{minipage}{0.95\linewidth}
%     % \vspace{3pt}
%     \centerline{\includegraphics[width=\textwidth]{figs/ablation_study.jpg}}
%     \end{minipage}
%     \caption{Ablation study on the MTIMNet model, divided by model structure, residual compensation, and multi-task loss. The checkmarks in each row indicate that corresponding modules are activated.}
% \end{figure}

\begin{table}[t]
\centering
\huge{
\caption{\rev{Ablation study of MTIMNet on the Customized Dataset. Lower is better for all metrics. The checkmarks indicates activated.}}
\label{tab_ablation_study}
\resizebox{\linewidth}{!}{
\begin{tabular}{l|ccc|cccc}
\hline
\multirow{2}{*}{Variant} & \multicolumn{3}{c|}{Modules} & \multicolumn{4}{c}{Error} \\
\cline{2-8}
& \makecell{Experts\\+Gates} & Residual & \makecell{Multi-task\\ Loss} & \makecell{ATE\\ (m)} & \makecell{RTE\\ (m)} & \makecell{PDE\\ (m)} & \makecell{AYE\\ (degree)} \\
\hline
MTIMNet (Ours)      & $\checkmark$ & $\checkmark$ & $\checkmark$ & \textbf{14.91} & \textbf{14.23} & \textbf{0.07}  & \textbf{2.03} \\
\hline
MLP-based          &             & $\checkmark$ & $\checkmark$ & 17.26 & 17.28 & 0.09  & 3.22 \\
LSTM-based         &             & $\checkmark$ & $\checkmark$ & 16.71 & 17.58 & 0.08  & 4.27 \\
TCN-based          &             & $\checkmark$ & $\checkmark$ & 18.88 & 19.14 & 0.09  & 4.50 \\
\hline
w/o Residual    & $\checkmark$ &             &             & 16.75 & 16.23 & 0.08  & 2.64    \\
w/o Uncertainty    & $\checkmark$ &   $\checkmark$   &             & 15.47 & 15.42 & \textbf{0.07}  & 2.14 \\
w/o Multi-task Loss& $\checkmark$ & $\checkmark$ &             & 16.25 & 16.00 & \textbf{0.07}  & 2.82 \\
w/o All            &             &             &             & 19.85 & 20.17 & 0.11 & 4.30 \\
\hline
\end{tabular}
}
}
\end{table}

\subsubsection{Model Structure}
We replace the experts and gating modules with multilayer perceptron (MLP)-, LSTM-, and TCN-based backbones, similar to IONet~\cite{Chen_Lu_Markham_Trigoni_2018}, RoNIN~\cite{9196860} and LLIO~\cite{wang2022llio}. 
\rev{The performance drop across all metrics, especially in trajectory and heading errors, shows that fixed shared features are insufficient, while the proposed expert-routing design provides more suitable task-specific motion representations.}

\subsubsection{Residual Estimation}
To validate the residual branch, we remove the residual estimation task and its loss term, while keeping the other components unchanged. 
\rev{Table~\ref{tab_ablation_study} shows consistent degradation across all metrics, with ATE increasing from $14.91 m$ to $16.75 m$ and AYE from 2.03 degrees to 2.64 degrees, confirming that the residual branch helps correct prediction bias and improve accuracy.}

\subsubsection{Uncertainty estimation}
\rev{To evaluate the effect of uncertainty modeling, we remove the uncertainty branch and its loss term, while preserving the residual branch and the expert-routing structure. 
Almost all metrics exhibited degradation, suggesting that uncertainty estimation is not redundant but rather provides useful, confidence-aware guidance.}

\subsubsection{Multi-task loss function}
To demonstrate the effectiveness of the loss function, we remove the residual estimation loss. 
Specifically, instead of learning residuals through independent tasks, we implicitly estimate the residuals in the model and compensate them to the corresponding tasks. 
As shown in the figure, the learning residuals implicitly result in an increase of 8\% for ATE and 28\% for AYE, respectively.

\noindent \textbf{Pseudo Wheel Speed Estimation:}
To evaluate the effectiveness of the anomaly detection module, we conducted ablation experiments in the fast-riding scenario by selectively removing the two components: Pearson Correlation Coefficient (PCC) and Comparison of Peak Counts (CPC). 
Table~\ref{tab_period_ablation} presents the results of the quantitative comparison.

\begin{table}[t]
  \centering
  \label{tab_period_ablation}
  \caption{Ablation study on pseudo wheel speed estimation.}
  \resizebox{1.0\linewidth}{!}{ % 将表格缩放到栏宽  
      \begin{tabular}{c|c|c|c|c|c}
        \hline
        Method & CEP68 & CEP80 & CEP95 & Max & TCR $\uparrow$ \\
        \hline
        w/ All    & \textbf{0.207}         & \textbf{0.274}         & \textbf{0.480}          & \textbf{0.723}        & 18.1\%        \\
        \hline
        w/o PCC         & 0.229         & 0.312         & 0.656         & 2.010         & 19.9\%        \\
        \hline
        w/o CPC         & 0.294         & 0.532         & 2.578         & 3.766        & 34.1\%        \\
        \hline
        w/o PCC \& CPC  & 0.331         & 0.796         & 2.540          & 3.766        & \textbf{37.4\%}        \\
        \hline
      \end{tabular}
  }
\end{table}

As shown, removing either component leads to an increase in error, while removing both causes the CEP95 to rise significantly from $0.48 m/s$ to $2.54 m/s$.
These results demonstrate the effectiveness of the anomaly detection module in filtering out abnormal data.

\subsection{Model Scale and Efficiency}

% old
% To analyze the number of parameters, the consumption of inference time, and the cost of the MTIMNet model, we compare the performance of different models on a server device equipped with a NVIDIA GeForce RTX 3090. 
% Table~\ref{tab_parameters_flops} presents the comparison results, showing that the MTIMNet model completes one estimation in 225.1 milliseconds under the CPU mode and 31.3 milliseconds under the GPU mode.
% Furthermore, compared with other models, MTIMNet possesses fewer parameters with less computational cost and achieves higher performance. 

To evaluate model complexity and inference efficiency, we compare MTIMNet with several baselines in terms of trainable parameters, inference time, and floating point operations (FLOPs) on a server device equipped with an NVIDIA GeForce RTX 3090, as shown in Table~\ref{tab_parameters_flops}. 
\rev{The MTIMNet has $1.085 M$ parameters and $3.759 M$ FLOPs, lower than most baselines and close to the lightweight recent method AirIO.
It also achieves the fastest CPU inference time ($4.0 ms$, tied with AirIO) and competitive GPU latency ($4.0 ms$), demonstrating a favorable trade-off between complexity, efficiency, and localization performance.}

\begin{table}[t]
    \centering
    \caption{Comparison of model complexity and inference time.}
    \label{tab_parameters_flops}
    \begin{tabular}{c|c|c|c}
    \hline
    Model & \makecell{Trainable\\ Parameters} $\downarrow$ & \makecell{Inference Time\\ (CPU / GPU)} $\downarrow$ & FLOPs $\downarrow$ \\
    \hline
    IONet   & \textbf{830.008 K} & 52.5 ms / 7.5 ms  & 33.830 M \\
    RoNIN   & 4.438 M   & 9.0 ms / 5.5 ms   & 21.225 M \\
    LLIO    & 7.158 M   & 8.5 ms / 6.5 ms   & 26.310 M \\
    UniTS   & 2.451 M   & 27.5 ms / 13.0 ms & 20.155 M \\
    IMUNet  & 4.068 M   & 202.5 ms / 7.0 ms & 97.733 M \\
    EqNIO   & 4.361 M   & 43.0 ms / 12.5 ms & 50.785 M \\
    AirIO   & 897.186 K & \textbf{4.0 ms} / \textbf{2.5 ms}   & 3.759 M \\
    \hline
    MTIMNet & 1.085 M   & \textbf{4.0 ms} / 4.0 ms   & \textbf{3.630 M} \\
    \hline
    \end{tabular}
\end{table}

% old
% \begin{table}[t]
%   \centering
%   \label{tab_parameters_flops}%
%   \caption{Performance comparison on parameters, inference time, and Floating Point Operations (FLOPs).}
%     \begin{tabular}{c|c|c|c}
%     \hline
%     Model & \multicolumn{1}{c|}{\makecell{Trainable\\ Parameters}} & \multicolumn{1}{c|}{\makecell{Inference Time\\ (CPU / GPU)}} & FLOPs \\
%     \hline
%     RoNIN & 4.438M & 276.1 ms / \textbf{21.6} ms & 21.225M \\
%     LLIO  & 7.158M & \textbf{222.0 ms} / 30.6 ms & 26.310M \\
%     UniTS & 2.451M & 381.0 ms / 25.4 ms & 20.155M \\
%     RF-Net & 13.040M & 451.3 ms / 43.9 ms & 1.172G \\
%     \textbf{MTIMNet} & \textbf{301.388K} & 225.1 ms / 31.3 ms & \textbf{409.248K} \\
%     \hline
%     \end{tabular}%
% \end{table}%

\subsection{Fusion Study}

To validate the effectiveness of pseudo wheel speed (PWS) for drift calibration, we run MTIMNet on \rev{free riding} trajectories and estimate PWS by continuously detecting valid pedalling periodicity. 
\rev{Once available, the PWS is converted into a displacement over the corresponding interval and fused with the MTIMNet displacement estimate using inverse-variance weighting.}
\rev{Table~\ref{tab_fusion} compares three settings: without PWS, with equal-weight fusion, and with inverse-variance weighted fusion, where the MTIMNet variance is predicted by the model and the PWS variance is derived from the corresponding error standard deviation.}
\rev{The results show that PWS improves localization accuracy, and inverse-variance weighted fusion performs best, reducing the ATE from $13.50 m$ to $10.81 m$.}
This demonstrates that PWS provides an effective auxiliary observation for reducing long-term drift.

\begin{table}[t]
\centering
\caption{\rev{Quantitative comparison of MTIMNet with different pseudo wheel speed fusion strategies on free riding trajectories.}}
\label{tab_fusion}
\begin{tabular}{c|c|c|c|c}
\hline
 & ATE & RTE & PDE & AYE \\
\hline
w/o pseudo wheel speed & 13.5  & 13.83 & \textbf{0.06} & 1.86 \\
\hline
\makecell{w/ pseudo wheel speed\\ (equal-weight fusion)} & 11.52 & 11.18 & \textbf{0.06} & 1.86 \\
\hline
\makecell{w/ pseudo wheel speed\\ (inverse-variance\\ weighted fusion)} & \textbf{10.81} & \textbf{10.35} & \textbf{0.06} & 1.86 \\
\hline
\end{tabular}
\end{table}

% old
% To validate whether the pseudo wheel speed is effective for speed calibration, we run the MTIMNet model on multiple trajectories while continuously detecting whether the rider's pedalling behaviour exhibits periodicity, thus estimating the wheel speed.
% Specifically, we compute the displacement from the available pseudo wheel speed to replace the one estimated by the MTIMNet model.
% We compared the performance of the MTIMNet model with and without pseudo wheel speed estimation, as shown in Table \ref{tab_fusion}.
% We observed that after calibration with the pseudo wheel speed, the ATE reduces from $12.59m$ to $10.35$ , with a 17.79\% improvement in accuracy.
% Furthermore, the accuracy of wheel speed estimation is independent of time length and does not impact the bike heading estimation. 
% Thus, the pseudo wheel speed enables effective compensation for the estimation errors accumulated by the MTIMNet model during the tracking process.

% 
% \begin{table}[t]
%   \centering
%   \label{tab_fusion}%
%   \caption{Quantitative comparison of the MTIMNet model with and without pseudo wheel speed calibration.}
%   \resizebox{1.0\linewidth}{!}{ % 将表格缩放到栏宽  
%     \begin{tabular}{c|c|c|c|c}
%     \hline
%           & ATE   & RTE   & PDE   & AYE \\
%     \hline
%     w/o Pseudo Wheel Speed & 12.59 & 12.32 & 0.06  & 1.99 \\
%     \hline
%     w/ Pseudo Wheel Speed & 10.35 & 10.46 & 0.06  & 1.99 \\
%     \hline
%     \end{tabular}%
%   }
% \end{table}%

\section{Conclusion}\label{sec:conclusion}

This paper proposes an inertial-only bike tracking framework in environments where GNSS signals are blocked or unavailable. It employs an inertial motion estimation model based on the mixture-of-experts structure to infer the bike's trajectory, and estimates wheel speeds from the rider's periodic pedalling pattern to calibrate the estimates in real time.
%For example, the model's uncertainty estimates and map priors could be leveraged to further enhance the accuracy and efficiency of locking shared bikes in different urban environments.
%Another future research direction is to introduce opportunistic signals (such as Wi-Fi, 5G, or road grid constraints) for adaptive fusion when available to improve tracking accuracy.
This enables large-scale GNSS-free device tracking and management for ride-hailing platforms.
Future work will further investigate model distillation strategies to support deployment on embedded devices, 
\rev{integrate opportunistic signals such as Wi-Fi and 5G for additional calibration, and explore collaborative localization among multiple shared bikes to enhance localization performance in more complex urban environments.}

%Future work may investigate model distillation strategies to enable deployment and inference on embedded devices, and integrate map with opportunistic signals (such as Wi-Fi and 5G) for calibration.

\bibliographystyle{IEEEtran}
\bibliography{tex/bibliography}

\begin{IEEEbiography}[{\includegraphics[width=1in,height=1.25in,clip,keepaspectratio]{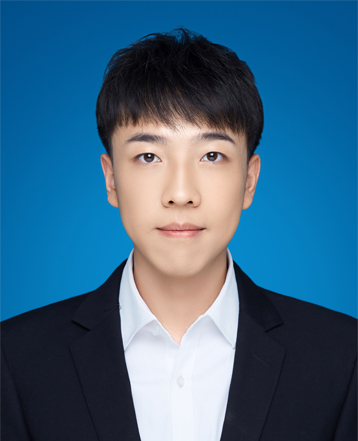}}]{\textbf{Feng~Liu}} received the B.S. degree from the Inner Mongolia University of Technology, Inner Mongolia, China, in 2022. He is currently working towards the Ph.D. degree in software engineering at Beijing Jiaotong University, China.
His research interests include multi-target positioning and mobile computing.
\end{IEEEbiography}

\begin{IEEEbiography}[{\includegraphics[width=1in,height=1.25in,clip,keepaspectratio]{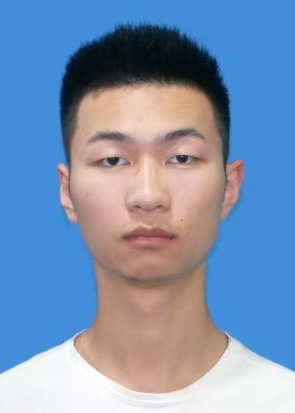}}]{\textbf{Kejia~Li}} received the B.S. degree in software engineering from Beijing Jiaotong University, Beijing, China, in 2024. He is currently pursuing the integrated M.S. and Ph.D. degree in the School of Cyberspace Security at Beijing Jiaotong University, Beijing, China.
His research interests include filtering algorithms, inertial navigation, and mobile computing.
\end{IEEEbiography}

\begin{IEEEbiography}[{\includegraphics[width=1in,height=1.25in,clip,keepaspectratio]{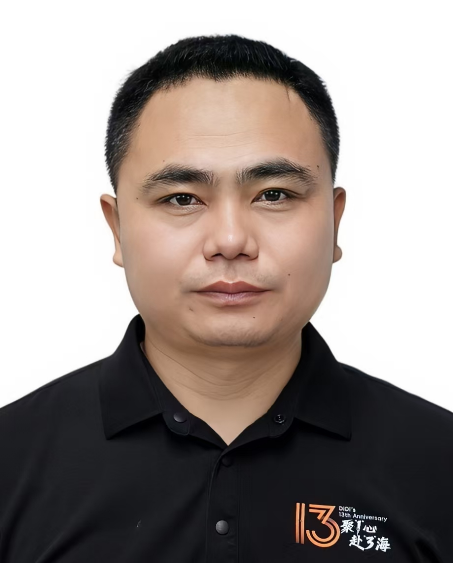}}]{\textbf{Zhiwei~Yang}} received his Bachelor of Science degree in computer science from Liaoning Petrochemical University, China, in 2005. He currently serves as an Embedded Systems Expert in the Two-Wheeler Sharing Department at DiDi. His research interests focus on positioning and sensor technologies, particularly their application to low-cost devices.
\end{IEEEbiography}

\begin{IEEEbiography}[{\includegraphics[width=1in,height=1.25in,clip,keepaspectratio]{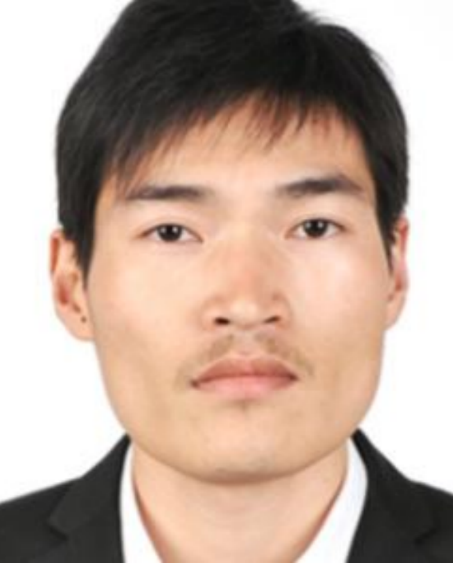}}]{\textbf{Chunwei~Yang}} received the MS degree from Zhejiang University, China, in 2016. He is currently the embedded software expert with the Hardware R\&D Department, DiDi Company. His research interests include positioning, machine learning, and deep learning.
\end{IEEEbiography}

\begin{IEEEbiography}[{\includegraphics[width=1in,height=1.25in,clip,keepaspectratio]{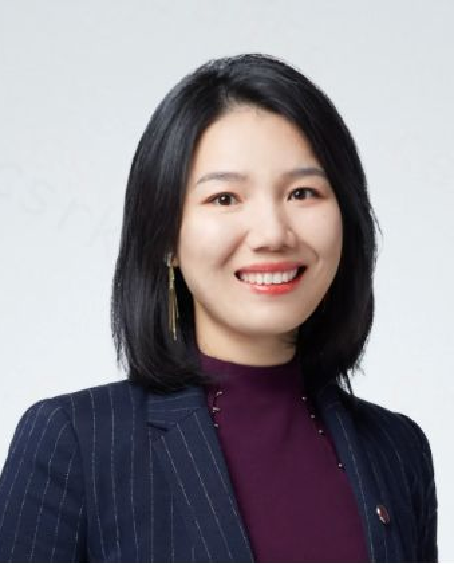}}]{\textbf{Qun~Li}} is the Head of DiDi Research Outreach, is leading DiDi’s efforts to work together with academic institutions worldwide, including research collaboration, talent cultivation and academic exchange. She was in the organizing committee of several workshops and tutorials. Her research interests lie in the intersection of artificial intelligence, autonomous driving and computer vision.
\end{IEEEbiography}

\begin{IEEEbiography}[{\includegraphics[width=1in,height=1.25in,clip,keepaspectratio]{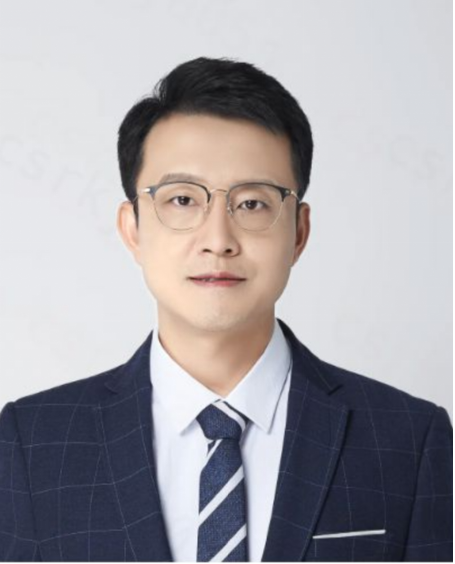}}]{\textbf{Guobin~Wu}} is the Director of Technology Ecology and Development Department at DiDi. His research interests include artificial intelligence and computer vision.
\end{IEEEbiography}
% \vspace{-15.5cm}

\begin{IEEEbiography}[{\includegraphics[width=1in,height=1.25in,clip,keepaspectratio]{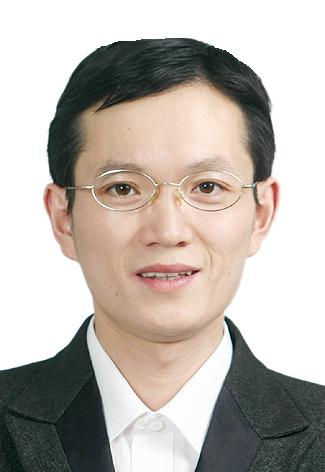}}]{\textbf{Qiang~Ni}} is currently a Professor and the Head of the Communication Systems Group, School of Computing and Communications, Lancaster University, Lancaster, U.K. His research interests include the area of future generation communications, computing and networking, including green communications and mobile networking, 5G and 6G, SDN, cloud networks, mobile computing, IoTs, cyber physical systems, AI, machine learning, big data analytics, and vehicular networks.
\end{IEEEbiography}

\begin{IEEEbiography}[{\includegraphics[width=1in,height=1.25in,clip,keepaspectratio]{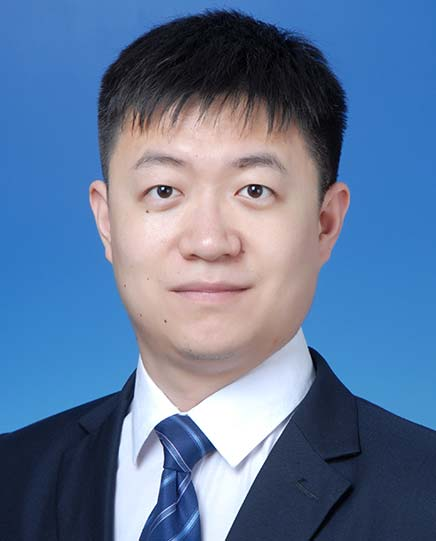}}]{\textbf{Ruipeng~Gao}} received the B.S. degree from the Beijing University of Posts and Telecommunications, China, in 2010, and the Ph.D. degree from Peking University, China, in 2016.
He was a Visiting Scholar with Purdue University, USA, in 2019. He is currently a Professor with the School of Cyberspace Science and Technology, Beijing Jiaotong University, China. His research interests include mobile computing and applications.
\end{IEEEbiography}

\newpage

% \section{Biography Section}
% If you have an EPS/PDF photo (graphicx package needed), extra braces are
%  needed around the contents of the optional argument to biography to prevent
%  the LaTeX parser from getting confused when it sees the complicated
%  $\backslash${\tt{includegraphics}} command within an optional argument. (You can create
%  your own custom macro containing the $\backslash${\tt{includegraphics}} command to make things
%  simpler here.)
 
\vspace{11pt}

% \bf{If you include a photo:}\vspace{-33pt}
% \begin{IEEEbiography}[{\includegraphics[width=1in,height=1.25in,clip,keepaspectratio]{fig1}}]{Michael Shell}
% Use $\backslash${\tt{begin\{IEEEbiography\}}} and then for the 1st argument use $\backslash${\tt{includegraphics}} to declare and link the author photo.
% Use the author name as the 3rd argument followed by the biography text.
% \end{IEEEbiography}

% \vspace{11pt}

% \bf{If you will not include a photo:}\vspace{-33pt}
% \begin{IEEEbiographynophoto}{John Doe}
% Use $\backslash${\tt{begin\{IEEEbiographynophoto\}}} and the author name as the argument followed by the biography text.
% \end{IEEEbiographynophoto}

\vfill

\end{document}